\documentclass{sig-alternate}
\usepackage{epsfig,latexsym,natbib}

\usepackage[utf8x]{inputenc}
\usepackage{amsmath}
\usepackage{mathtools}
\DeclarePairedDelimiter{\abs}{\lvert}{\rvert}

\usepackage[breaklinks]{hyperref}
\hypersetup{colorlinks=true,citecolor=blue,linkcolor=blue}

\usepackage{algorithm}
\usepackage[noend]{algorithmic}
\usepackage{multirow}
\usepackage{color}
\usepackage{subfigure}
\usepackage{xspace}
\usepackage{booktabs}

\usepackage{url}

\usepackage{soul}

\begin{document}

%Define Algorithm Names
\newcommand{\mlaslcs}{\textsc{MlS}-LCS\xspace}
\newcommand{\scs}{\%[SCS]\xspace}
\newcommand{\rakel}{RAkEL\xspace}
\newcommand{\mlknn}{\textsc{Ml}kNN\xspace}
\newcommand{\toy}{\textsc{toy$6$x$4$}\xspace}
\newcommand{\pos}{\textsc{mlposition$4$}\xspace}
\newcommand{\posN}{\textsc{mlpositionN}\xspace}

%\ecjHeader{x}{x}{xxx-xxx}{201X}{Inducing Generalized Multi-Label Rules with LCS}{F. A. Tzima, M. Allamanis, A. Filotheou, and P. A. Mitkas}
\title{Inducing Generalized Multi-Label Rules with Learning Classifier Systems}  

\author{%
Fani~A.~Tzima${}^\ddag$
\quad Miltiadis Allamanis$^{\dag}$\thanks{Work done primarily while the author was an undergraduate student at the Aristotle University of Thessaloniki}
\quad Alexandros Filotheou${}^\ddag$
\quad Pericles~A.~Mitkas${}^\ddag$ \and%
\begin{tabular}{cc}
\begin{tabular}{c}
$^{\ddag}$\affaddr{Department of Electrical and Computer Engineering} \\
\affaddr{Aristotle University of Thessaloniki (AUTh)} \\
\affaddr{Thessaloniki, 53624, Greece} \\
\email{fani@issel.ee.auth.gr} \\\email{alexandros.filotheou@gmail.com} \\\email{mitkas@eng.auth.gr}
\end{tabular}
&
\begin{tabular}{c}
$^{\dag}$\affaddr{School of Informatics} \\
\affaddr{University of Edinburgh} \\
\affaddr{Edinburgh, EH8 9AB, UK} \\
\email{m.allamanis@ed.ac.uk}
\end{tabular}
\end{tabular}
\setlength{\tabcolsep}{6pt}
}% end author

\date{}
\maketitle

\begin{abstract}
In recent years, multi-label classification has attracted a significant body of research, motivated by real-life applications, such as text classification and medical diagnoses. Although sparsely studied in this context, Learning Classifier Systems are naturally well-suited to multi-label classification problems, whose search space typically involves multiple highly specific niches. This is the motivation behind our current work that introduces a generalized multi-label rule format -- allowing for flexible label-dependency modeling, with no need for explicit knowledge of which correlations to search for -- and uses it as a guide for further adapting the general Michigan-style supervised Learning Classifier System framework. The integration of the aforementioned rule format and framework adaptations results in a novel algorithm for multi-label classification whose behavior is studied through a set of properly defined artificial problems. The proposed algorithm is also thoroughly evaluated on a set of multi-label datasets and found competitive to other state-of-the-art multi-label classification methods.
\end{abstract}

%\begin{keywords}
%multi-label classification, learning classifier systems, algorithm adaptation
%\end{keywords}

\section{Introduction}
Every day massive amounts of data are collected and processed by computers and embedded devices. 
This data, however, is useless to the people and organizations collecting it, unless it can be properly processed and converted into actionable knowledge. Machine Learning (ML) \citep{murphy2012} techniques 
are especially useful in such domains, where  automatic extraction of knowledge from data is required. 

One of the most common and extensively studied knowledge extraction tasks is classification.
In traditional classification problems, data samples are associated with a single category, termed \emph{class}, that may have two or more possible values. For example, the outlook for tomorrow's weather may be `sunny', `overcast' or `rainy'. On the other hand, \emph{multi-label classification}\footnote{
Multi-label classification can be viewed as a particular case of the \emph{multi-dimensional problem} \citep{Read2013}  where the goal is to assign each data sample to multiple multi-valued (in contrast to binary, for the multi-label case) classes. 
}, that is the focus of our current investigation, involves problems where each sample is associated with one or more binary categories, termed \emph{labels}. 
For example, a newspaper article about climate change can be described by both tags `environment' and `politics'; or a patient can simultaneously be diagnosed with `high blood pressure', `diabetes' and `myopia'.

Although single-label classification problems have been thoroughly explored, with the aid of various ML algorithms, 
literature on multi-label classification is far less abundant. Multi-label classification problems are, however, by no means less natural or intuitive and are, in fact, very common in real-life. 
The fact that, until recently, only a few of the corresponding problems were tackled as multi-label is mainly due to computational limitations. Recent research (see \citet{tsoumakas2010} and \citet{read2010scalable} for overviews) and modern hardware, though, has made multi-label classification more affordable. A gradually increasing number of problems are now being tackled as multi-label, allowing for richer and more accurate knowledge mining in real-world domains, such as medical diagnoses, protein function prediction and semantic scene processing. 

A careful inspection of the corresponding literature reveals that multi-label classification is nowadays a widely popularized task, but Evolutionary Computation (EC) approaches to prediction model induction are very sparse. The few approaches that exist \citep{vallim2008new, vallim2009multi, abhari2011} explore the use of Michigan-style Learning Classifier Systems (LCS) \citep{holland} -- a Genetics-based ML %\citep{kovacs2009genetics}  
method that combines EC and reinforcement \citep{XCS} or supervised learning \citep{bernado2003,orriols08b} -- but report promising results only on small artificial and real-world problems. Although they lack an extensive experimental evaluation, however, both in terms of target multi-label classification problem and rival algorithm variety, they are still based on a valid premise. This premise actually also summarizes the motivation of our current work:  LCS, due to their inherent characteristics, are naturally suited to multi-label classification and can provide an effective alternative in problem domains where highly expressive human-readable knowledge needs to be extracted, while maintaining low inference complexity.

Indeed, in recent years, LCS have been modified for data mining \citep{DBLP:series/sci/2008-125} and single-step classification problems, notably in the UCS \citep{orriols08b} and the SS-LCS frameworks \citep{tzima2012, tzima2013}. Their niche-based update and their overall iterative (rather than batch) learning approach has been shown to be very efficient in domains where different problem niches occur (including multi-class and unbalanced classification problems). Thus, we believe that it will also allow them to tackle the multiple and often very specific niches that comprise the search space of multi-label classification problems. 

Moreover, LCS may provide a practical alternative to deterministic methods, when exhaustive search is intractable (for example, in multi-label classification problems with large numbers of labels and/or attributes) or, in general, when targeting problems with large, complex and diverse search spaces. In such cases, the global search capability of EC, combined with the local search ability of reinforcement learning, allows LCS to evolve flexible, distributed solutions, wherein discovered patterns are spread over a population of (individual or groups of) rules, each modeling a niche of the problem space \citep{Urbanowicz2009}. 

LCS are also model-free and, thus, do not make any assumptions about target data (e.g. number, types and dependencies among attributes, missing data, distribution of training instances in terms of the target categories). This allows them to identify all kinds of relationships -- including epistatic ones that are characteristic of multi-label domains -- both between the feature and label space and among the various labels. 

Finally, as already mentioned, the nature of the knowledge representation evolved by LCS is a great advantage in certain application domains, where rule comprehensibility is an important requirement. At this point is should be noted that, Michigan-style LCS, although implicitly geared towards maximally accurate and general rules, tend to evolve rather large populations, mainly due to the distributed nature of the evolved solutions and the retainment of inexperienced rules created by the system's exploration component. Ruleset compaction techniques are, though, available to reduce the number of rules in the final models and enhance their comprehensibility. 

Overall, the aim of our current work (that builds on previous research presented in \citet{allamanis13}) is to develop an effective LCS algorithm for multi-label classification.
In this direction, we employ a general supervised learning framework and extend it, to render it directly applicable to the corresponding problems, without the need for any problem transformation. More specifically, we adapt three major components of the traditional LCS architecture: 
(i) ~the \emph{Rule Representation}, to allow for rule consequents that include multiple labels;
(ii)~the \emph{Update Component}, to consider multiple correct labels in rule parameter updates; and
(iii)~the \emph{Performance Component}, to enable inference in multi-label settings where multiple concurrent
 decisions are required.

The aforementioned extensions implicitly define the structure and main contributions of the paper, which are detailed
 after briefly presenting the relevant background (Section \ref{sec:background}).
Briefly, our current work's main contributions are: 
\begin{itemize}
\item a \emph{generalized multi-label rule format} (Section \ref{sec:rules4ml}) that has several distinct advantages over
 those used in other multi-label classification methods; 
\item a \emph{multi-label Learning Classifier System} (Section \ref{sec:mlLCS}), named the Multi-Label Supervised Learning Classifier System (\mlaslcs), whose components allow
 for efficient and accurate multi-label classification through developing expressive multi-label rulesets; and 
\item an experimental evaluation (Section \ref{sec:evaluation}) of our proposed LCS approach, against other
 state-of-the-art algorithms on widely used datasets, that validates its potential.
\end{itemize}
Section \ref{sec:concl} restates our overall contributions, outlines future research directions and concludes this work with additional insights on the potential of the proposed algorithm.

\section{Background}
\label{sec:background}
\subsection{Multi-label Classification}
Multi-label classification is a generalization of traditional classification where each sample is associated
 with a set of mutually non-exclusive binary categories, or \emph{labels}, $Y \subseteq L$. Thus, defining the problem from a machine
 learning point of view, a multi-label classification model approximates a function $f$:~$X \rightarrow L^*$ where $X$ 
 is the feature space and $L^*$ is the powerset of the label space $L$ (i.e., the powerset of the set of all possible labels). 
 
The general multi-label classification framework, by definition, implies the existence of an additional dimension:
 that of the multiple labels which data samples can be associated with. This additional complexity affects not only
 the learning processes that can be applied to the corresponding problems, but also the procedures employed during the evaluation of developed models \citep{tsoumakas2010}. 

The basic premise that differentiates learning, with respect to the single-class case, is that to provide more accurate predictions, label correlations should be factored in multi-label classification models. This need is based on the observation that labels occur together with different frequencies. For example, a newspaper article is far more likely to be assigned the pair of tags `science' and `environment', than the pair `environment' and `sports'. 
 Of course, in the absence of label correlations, the corresponding multi-label problem is trivial and can be completely broken down (without any loss of useful information) to  $|L|$ binary decision problems.
 
There are three main approaches to tackling multi-label classification problems in the literature: problem transformation, algorithm transformation (such as the LCS approach presented in this paper) and ensemble methods.%tsoumakas2007multi}:

\emph{Problem Transformation} methods transform a multi-label classification problem into a set of single-label ones. Various such transformations have been proposed, involving different trade-offs between training time and label correlation representation. The simplest of all transformations is the Binary Relevance (BR) method \citep{tsoumakas2007multi}, to which the Classifier Chains (CC) method \citep{read2009classifier} is closely related. Other transformations found in the literature are Ranking by Pairwise Comparison (RPC) \citep{hullermeier2008} and the Label Powerset (LP) method that has been the focus of several studies, including the Pruned Problem Transformation (PPT) \citep{read2008} and HOMER \citep{tsoumakasHOMER2008}. 

 \emph{Algorithm Transformation} methods adapt learning algorithms to directly handle multi-label data.
Such methods include: (a) several multi-label variants (\mlknn) of the popular $k$-Nearest Neighbors lazy learning algorithm \citep{zhang2007ml}, %, mlknn2}, 
as well as hybrid methods combining logistic regression and  $k$-Nearest Neighbors \citep{cheng09}; (b)~multi-label decision trees, such as ML-C4.5 \citep{clare01} and predictive clustering trees (PCTs) %\citep{blockeel98,vens2008decision}; 
\citep{vens2008decision}; 
(c)~Adaboost.MH and Adaboost.MR \citep{schapire2000boostexter}, that are two extensions of Adaboost.MH for multi-label learning; (d) several neural network approaches \citep{crammer03, zhang06}; (e) the Bayesian Networks approach by \citet{zhang2010multi}; (f) the 
SVM-based ranking approach by \citet{elisseeff05}; and (g) the associative classification approach of MMAC \citep{thabtah2004mmac}.

 \emph{Ensemble} methods are developed on top of methods of the two previous categories. The three most well-known ensemble methods employing problem transformations as their base classifiers are \rakel\citep{tsoumakas2011}, ensembles of pruned sets (EPS) \citep{read2008multi} and ensembles of classifier chains (ECC) \citep{read2009classifier}. On the other hand, an example of an ensemble method where the base classifier is an algorithm adaptation method (i.e., provides multi-label predictions) can be found in %\citep{kocev07,kocev07thesis} 
 \citet{kocev07thesis} where ensembles of predictive clustering trees (PCTs) are presented.

As far as the evaluation of multi-label classifiers is concerned, several traditional evaluation metrics can be used, provided that they are properly modified.  The specific metrics employed in our current study for algorithm comparisons are \emph{Accuracy}, \emph{Exact Match (Subset Accuracy)} and \emph{Hamming Loss}. In what follows, these metrics are defined for a dataset $D$, consisting of multi-label instances of the form $(x_i,Y_i)$, where $i=1\dots|D|$, $Y_i \subseteq L$ ($Y_i \in L^*$), $L$ is the set of all possible labels and $\hat{Y}_i=H(x_i)$ is a prediction function.

\emph{Accuracy} is defined as the mean, over all instances, ratio of the size of the intersection and union sets of actual
 and predicted labels. It is, thus, a label-set-based metric, defined as: 
\begin{equation} \label{eq:acc} \textsc{Accuracy}(H,D)=\frac{1}{|D|} \sum_{i=1}^{|D|} \frac{|Y_i \bigcap \hat{Y}_i|}{|Y_i \bigcup \hat{Y}_i|} 
\end{equation}

 \emph{Exact Match (Subset Accuracy)} is a simple and relatively strict evaluation metric, calculated 
  as the label-set-based accuracy: 
  \begin{equation} \textsc{Exact-Match}(H,D)=\frac{|C|}{|D|} 
  \end{equation}  
  where $C$ is the set of correctly classified instances for which  $Y_i \equiv \hat{Y}_i$.

\emph{Hamming Loss} corresponds to the label-based accuracy, taking into account false positive and false negative predictions and is defined as: 
 \begin{equation} 
\textsc{Hamming-Loss}(H,D)=\frac{1}{|D|} \sum_{i=1}^{|D|} \frac{\hat{Y}_i \varDelta Y_i }{|L|} 
 \end{equation}  
 where $Y_i \varDelta \hat{Y}_i$ is the symmetrical difference (logical XOR) between $\hat{Y}_i$ and $Y_i$.

The interested reader can find an extensive discussion on the merits and trade-offs of various multi-label classification methods and evaluation measures, along with the latter's definitions, 
in \citet{tsoumakas2010, read2010scalable, madjarov2012}.

\subsection{Learning Classifier Systems}
Learning Classifier Systems (LCS) \citep{holland} are an evolutionary approach to supervised and reinforcement learning problems. 
Several flavors of LCS exist in the literature \citep{Urbanowicz2009}, with most of them following the ``Michigan approach'', such as (a) the strength-based ZCS \citep{WilsonZCS, TzimaZCS} and SB-XCS \citep{kovacs02, kovacs02a}; and (b)~the accuracy-based XCS \citep{XCS} and UCS \citep{bernado2003, orriols08b}. 
Accuracy-based systems have been the most popular so far for solving a wide range of problem types \citep{DBLP:series/sci/2008-125} -- such as classification \citep{butzXCS2, orriols09, fernandez2010}, regression \citep{XCSF, Butz2008, Stalph2012}, sequential decision making \citep{Butz2005, Lanzi2006b}, and sequence labeling \citep{nakata2014, nakata2015} --  in a wide range of application domains -- such as medical diagnoses \citep{kharbat07}, 
fraud detection \citep{Behdad2012} and robot arm control \citep{Kneissler2014}.

Given that multi-label classification is a supervised task, we chose to tackle the corresponding problems using supervised (Michigan-style) LCS. Such LCS maintain a cooperative population of condition-decision rules, termed classifiers, and combine supervised learning %\citep{murphy2012} 
with a genetic algorithm (GA). The GA works on classifier conditions in an effort to adequately decompose the target problem into a set of subproblems, while supervised learning evaluates classifiers in each of them \citep{Lanzi}. 
The most prominent example of this class of systems is the accuracy-based UCS algorithm \citep{bernado2003, orriols08b}. Additionally, we have recently introduced SS-LCS, a supervised strength-based LCS, that 
provides an efficient and robust alternative for offline classification tasks \citep{tzima2012, tzima2013} by extending previous strength-based frameworks 
\citep{WilsonZCS, kovacs02, kovacs02a}.

To sum up, our current investigation focuses on developing a supervised accuracy-based Michigan-style LCS for multi-label classification by extending the base 
architecture of UCS and incorporating the clustering-based initialization component of SS-LCS. It also builds on our research presented in \citet{allamanis13}, from which the main differences are: (a)~the multi-label crossover operator (Section~\ref{sec:disc_comp}); (b)~the modified deletion scheme and the population control strategy (Section~\ref{sec:pop_control}); (c)~the clustering-based initialization process (Section~\ref{sec:init}); and, more importantly, (d)~the extensive experimental investigation of the proposed algorithm, both in terms of target problems and rival algorithms (Section~\ref{sec:evaluation}). The last point also addresses the main shortcoming of existing multi-label LCS approaches \citep{vallim2008new, vallim2009multi, abhari2011}, namely the absence of empirical evidence on their potential for multi-label classification in real-world settings.

\subsection{Rule Representation in LCS}
LCS were initially designed with a ternary representation: rules involved conditions represented as fixed-length bitstrings  defined over the alphabet \{0, 1, \#\} and numeric actions. To deal with continuous attributes, often present in real-world classification problems, however, interval-based rule representations were later introduced, starting with Wilson's \emph{min-max representation} that codifies continuous attribute conditions using the lower $l_i$ and upper $u_i$ limit of the acceptable interval of values. When using this representation, invalid intervals (where $l_i$$>$$u_i$) -- and, thus, impossible conditions -- may be produced by the genetic operators. A simple approach to fixing this problem was proposed in \citet{Stone2003} that introduced the \emph{unordered-bound representation} -- the most popular representation used for continuous attributes in LCS in the last few years. The unordered-bound representation proposes the use of interval limits without explicitly specifying which is the upper  and which the lower bound: the smaller of the two limit values is considered to be the interval's lower bound, while the larger is the upper bound. The unordered-bound representation is our representation of choice for continuous attributes in our current work.

Other than interval-based ones, several other rule representations have been introduced for LCS (mainly XCS and UCS) during the last few years. These representation aim to enable LCS to deal with function approximation \citep{ XCSF} and real-world problems, and include hyper-elipsoidal representations \citep{Butz2008},  convex hulls \citep{Lanzi2006} and tile coding \citep{Lanzi2006b}. Other more general approaches used to codify rules are neural networks \citep{Bull2002}, messy representations \citep{lanzi1999a} and S-expressions \citep{lanzi1999b}, fuzzy representations \citep{orriols2009}, genetic-programming like encoding schemes involving code fragments in classifier conditions \citep{Iqbal2014}, and dynamical genetic programming \citep{preen2013}. %bull2009, preen2014}.

\section{Rules for Multi-label Classification}
\label{sec:rules4ml}

To tackle multi-label classification problems with rule-based methods, and thus also with LCS, we need an expressive rule format, able to capture correlations both between the feature and label space and among the various labels. In this Section, we introduce a rule format that possesses these properties and forms the basis of our proposed multi-label LCS, detailed in Section \ref{sec:mlLCS}. In the last part of the Section, we also describe some ``internal'' rule representation issues.

\subsection{Generalized Multi-label Rule Representation} \label{sec:genMLrules}
Single-label classification rules traditionally follow the production system (or  ``if-then'') form
 $r_i: condition_i \rightarrow y_i$,
where rule's $r_i$ condition comprises a conjunction of tests on attribute values and its consequent $y_i$ contains a single value from the target classification problem's set of possible categories (or classes). 
It is also worth noting that, for zero-order rules, the condition comprises $k$ ($0 \leq k \leq |X|$) tests 
$$(x_{1} \text{ op }  u_{1}) \wedge (x_{2} \text{ op } u_{2}) \wedge \dots \wedge (x_{k} \text{ op } u_{k})$$ 
wherein $x_i$ $\in$ $X$ is one of the problem's attributes, $op$ is an operator, and $u_i$ is a constant set, number or range of numbers. 

It is evident that 
%rules of this kind (i.e., 
rules following the form described above are not able to readily handle multi-label classifications. To alleviate this shortcoming, we introduce a modification to the rule consequent part, such that, for any given multi-label rule $r_i: condition_i \rightarrow Y_i$, the consequent part $Y_i$ takes the form: 
\begin{equation}
\label{eq:mlRule}
Y_i =  (l_{1}\in \{0,1\}) \wedge \dots \wedge (l_{m}\in \{0,1\})
\end{equation}
where $l_i$ is one of the problem's possible labels ($l_i \in L_{sub}$, $L_{sub} \subseteq L$), taking either the value $1$  for labels advocated by rule $r_i$, or the value $0$ in the opposite case.

According to Eq. \ref{eq:mlRule}, the consequent part of a rule following our proposed \emph{Generalized Multi-label Representation} includes both the labels $l_{ai}$ that the rule advocates for ($l_{ai}$=$1$, $i \in A$), and the labels $l_{oj}$ it is opposed to ($l_{oj}$=$0$, $j \in O$). 
It should be noted that (i)~no label can appear more than once in the rule consequent part ($A \cap O = \emptyset$) and (ii)~rules are allowed to ``not care'' about certain labels, which are, thus, absent from the rule consequent ($A \cup O = L_{sub}\subseteq L$). In other words, the proposed rule format has the important property of being able to \emph{map rule conditions to arbitrary subspaces of the label-space}.

An abbreviated notation for rule consequent parts can be derived by using the ternary alphabet and substituting ``1'' for advocated labels, ``0'' for labels the rule is opposed to and ``\#'' for ``don't cares''. Thus, in a problem with three labels, a rule advocating the first label, being indifferent about the second and opposed to the third is denoted as: $(condition) \rightarrow 1\#0$.

Rules following our proposed \emph{Generalized Multi-label Representation} have some unique properties. First, rules are easy to interpret, rendering the discovered knowledge (rulesets) equally usable by both humans and computers. Such a property is important in cases where providing useful insights to domain experts is amongst the modelers' goals. 

Furthermore, rules have a flexible label-correlation representation. Algorithms inducing \emph{generalized multi-label rules} do not require explicit knowledge of which label correlations to search for and can variably correlate the maximum possible number of labels to any given condition. Therefore, in contrast to problem transformation methods that need to explicitly create (at least) one model for each possible label correlation/combination being searched for, algorithms inducing generalized multi-label rules can approach all possible spectra between the BR (not looking into any label correlations) and LP (searching for all possible label combinations) transformations and simultaneously create the most compact rule representation of the problem-space, with no redundancy. 

Consider, for example, the (artificial) problem \toy with 6 binary attributes and 4 labels, where the first two labels only depend on the first two attributes, according to the rules\footnote{These are actually the rules, without default hierarchies, defining the artificial problem studied in \citep{vallim2009multi}.}
\begin{equation}\label{lala}
\begin{array} {c c}
1\#\#\#\#\#	\rightarrow 01\#\#  \hspace{5mm}	& 00\#\#\#\#  \rightarrow 11\#\# 	\\
 	\hspace{5mm}								& 01\#\#\#\#  \rightarrow 10\#\# \\
\end{array}
\end{equation}
and the last two labels always have exactly the same values as the last two attributes. The \emph{shortest complete solution (SCS)} (i.e., the solution containing the smallest possible number of rules that allow for specific decisions to be made for all labels of all data samples), given our generalized rule format, involves 7 rules: the 3 rules in Eq.~\ref{lala}, plus one of the following alternative rulesets.
\begin{equation*}
\begin{array}{c c}
\text{Ruleset A$_1$} \hspace{2mm} & \hspace{2mm}  \text{Ruleset A$_2$}\\
\overbrace{\#\#\#\#00  	\rightarrow \#\#00}  	\hspace{5mm}		& 
\overbrace{\#\#\#\#\#0  	\rightarrow \#\#\#0}	\\
\#\#\#\#01  	\rightarrow \#\#01 	\hspace{5mm}		& \#\#\#\#\#1  	\rightarrow \#\#\#1	\\
\#\#\#\#10  	\rightarrow \#\#10 	\hspace{5mm}		& \#\#\#\#0\#  	\rightarrow \#\#0\#	\\
\#\#\#\#11 	\rightarrow \#\#11	\hspace{5mm}		& \#\#\#\#1\#  	\rightarrow \#\#1\#
\end{array}
\end{equation*}

If we do not use the generalized rule format, we are bound to induce rules with all-specific consequents that are not allowed to ``don't care'' about any of the problem's labels. This would be equivalent to the LP transformation, creating rules for each possible label combination, and would result in (at least) 12\footnote{For this particular problem, we need 12 and not $|L^*|$=16 rules, as some label combinations are missing from the training dataset and, thus, no model would need to be built for them.} rules for our current example -- i.e., the combinations of each of the first 3 rules with each of the 4 rules in set A$_1$.

\subsection{Rule Representation in Chromosomes}
Rules in \mlaslcs, not unlike traditional LCS, are mapped to chromosomes -- consisting of 1s and 0s -- to be used in the GA. Our approach 
universally employs an \emph{activation bit}, indicating whether a test for a specific attribute's values is active or inactive (\#), irrespective of the attribute type. Thus,
 \emph{binary attributes} and \emph{labels} are represented using two bits. %$b_i$. 
 The first bit represents the activation status of the corresponding test and the second bit represents the target (attribute or label) value.
 \emph{Nominal attributes} are represented by $n+1$ bits, where $n$ is the number of the attribute's possible values. 
For \emph{continuous attributes} we employ the ``unordered-bound representation'' \citep{Stone2003}, defining an acceptable range of values for an attribute $x_i$ through two bounds $b_1$ and $b_2$, such that $min(b_1,b_2) \leq x_i \leq max(b_1,b_2) $. The two threshold values $b_1$ and $b_2$ are represented by binary numbers discretized in the range $[x_{min}, x_{max}]$ where $x_{min}$ ($x_{max}$) is the lowest (highest) possible value for attribute $x_i$. The number of bits used in this representation is $2k+1$, where $k$ determines the quantization levels ($2^k$) and the additional bit is the activation bit.

\section{LCS for Multi-label Classification}
\label{sec:mlLCS}
As already mentioned, the scope of our current work comprises offline multi-label classification
problems -- that is classification problems that can be described by a collection of data and do not involve online interactions. We tackle these problems using \emph{Michigan-style supervised LCS}. 

Such LCS have been successfully used for evolving rulesets in single-label classification domains. In these cases, evolved  
rulesets $R$ comprise \emph{cooperative} rules that collectively solve the target problem, while they are also required to be maximally \emph{compact}, i.e., containing the minimum number of rules that are necessary for solving the problem. Equivalently, all rules $r_i \in R$ need to have \emph{maximally general conditions}, that is the greatest possible feature space coverage. 
Additionally,
a ruleset $R$ is considered an effective solution if it contains rules that are adequately \emph{correct}, with respect to a specific performance/correctness metric.

While all the aforementioned properties are also desirable in generalized multi-label rulesets (i.e., rulesets comprising generalized multi-label rules, as described in Section \ref{sec:genMLrules}), there is  
an additional important requirement. These rulesets also need to \emph{exhaustively cover the label space}. In other words, rules in a multi-label ruleset $R$ should collectively be able to decide about all labels for every instance. This latter desirable property, together with the compactness requirement, indicates that multi-label rules should ideally have maximally \emph{general} conditions and combine them with the corresponding maximally \emph{specific} consequents.

Consider, for example the following two rules for the \toy problem:
\begin{equation} \label{eq:rules}
\begin{array} {c c}
 r_1\text{: }1\#\#\#\#\#	\rightarrow 01\#\#  	& 
\hspace{-0.5mm} r_2\text{: }1\#\#\#\#\#  \rightarrow 0\#\#\# 	\\ 
\end{array}
\end{equation}
Both rules are perfectly accurate (for the labels for which they provide concrete decisions), but the first rule is clearly preferable, correlating the (common) condition with a larger part of the label space and, thus, promoting solution compactness.

Overall, it is evident that algorithms building rulesets for multi-label classification problems need to consider the trade-off between condition generalization, consequent specialization and rule correctness. In an LCS setting, this means that the core learning and performance procedures need to be appropriately modified to effectively cope with multi-label problems.
Thus, translating the aforementioned desirable properties of multi-label rulesets into concrete design choices towards formulating our proposed multi-label LCS algorithm, we derive the following requirements for its components: 
\begin{itemize}
\item
the \emph{Performance Component}, that is responsible for using the rules 
developed to classify previously unseen samples, needs to be modified to 
enable effective inference based on (generalized) multi-label rules;
\item the \emph{Update Component}, which is responsible for updating 
rule-specific parameters, such as accuracy and fitness, needs an 
appropriate \emph{metric of rule quality}, taking into account that generalized multi-label rules make decisions over subsets of labels that may, additionally, be only partially correct;
\item
the \emph{Discovery Component} that explores the search space and produces 
new rules through a \emph{steady-state GA}, needs to focus on evolving 
multi-label rulesets that are accurate, complete and cover both the 
feature and label space.\emph{Subsumption conditions}, controlling rule ``absorption'', also need to be adapted, so as to favor rules with more general conditions and more specific consequents.
\end{itemize}
These substantial adaptations to the general LCS framework, essentially define the 
proposed \mlaslcs algorithm and are presented in detail in the following Sections.

%------------------------------------------------------------- Description of the algorithm ------------------------------------------------------------- 
\subsection{The Training Cycle of the Multi-Label Supervised Learning Classifier System (\textsc{MlS-LCS}) }
\mlaslcs employs a population $\mathbf{P}$ of gradually evolving, cooperative classifiers (rules) that collectively form the solution to the target classification task, by each encoding a fraction (niche) of the problem domain. Associated with each classifier, there are a number of parameters: 
\begin{itemize}
 	\item the numerosity $num$, i.e., the number of classifier
 	copies (or microclassifiers) currently present in the ruleset; 
 	\item the correct set size $cs$ that estimates the average size of 
 	the correct sets the classifier has participated in; 
 	\item the time step $ts$ of the last occurrence of a GA in a correct 
 	set the classifier has belonged to; 
 	\item the experience $exp$ that is measured as the classifier's number
	 of appearances in match sets (multiplied by the number of labels); 
	 \item the effective match set appearances $msa$ that is the classifier's experience, (possibly) reduced by a certain amount for each label that the classifier did not provide a concrete decision for (see Eq.~\ref{eq:msa}); 
	 \item the number of the classifier's correct and incorrect label decisions,
 	$tp$ and $fp$ respectively; 
 	\item the accuracy  $acc$ that estimates the probability of a
 	 classifier predicting the correct label; and 
 	\item the fitness $F$ that is a measure of the classifier's quality.
\end{itemize}

At each discrete time-step $t$ during training, \mlaslcs receives a data instance's $V_t$ attribute values $X_t$ and labels $Y_t$ ($V_t: X_t \rightarrow Y_t$ \big| $ Y_t \subseteq L$) and follows a cycle of \emph{performance}, \emph{update} and \emph{discovery}  component activation (Alg.~\ref{alg:mlaslcs}). The completion of a full training cycle is followed by a new cycle based on the next available input instance, provided, of course, that the algorithm's termination conditions have not been met.

\begin{algorithm}
\caption{\mlaslcs component activation cycle during training (at step $t$).} \label{alg:mlaslcs}
\vspace{1mm}
\texttt{RUN\_TRAINING\_CYCLE } 
\begin{algorithmic}[1]
  	\STATE $V_t \gets$ read next data instance 
	\STATE initialize empty set $\mathbf{M}$
	\FOR {\textbf{each} label $l \in L$} 
		\STATE initialize empty sets $\mathbf{C}[\:l\:]$ and $\mathbf{!C}[\:l\:]$ 
	\ENDFOR
  	\STATE $\mathbf{M} \leftarrow$ generate match set out of $\mathbf{P}$ using $V_{t}$ 
  	\IF {deletions have commenced}\label{condition:rouletteWheelDeletionsCommenced}
  		\STATE control match set $\mathbf{M}$ %{\color{red}size considering $\mathbf{P}$} 
		\label{func:controlMatchSet}
  	\ENDIF
  		
	\FOR {\textbf{each} $l \in L$} 
		\STATE $l_n \gets labels(V_t)[\:l\:]$
		\STATE $\mathbf{C}[\:l\:] \gets$ generate label correct set out of $\mathbf{M}$ using $l_n$
		\STATE $\mathbf{!C}[\:l\:] \gets$ generate label incorrect set out of $\mathbf{M}$ using $l_n$
	\ENDFOR
	
	\FOR {\textbf{each} classifier $cl \in \mathbf{M}$}
		\STATE \texttt{UPDATE\_FITNESS} ( $cl$ ) %{\color{red}considering $\mathbf{C}$ }
		\IF{$\exists l_{i} \in L$ such that $cl \in \mathbf{C}[\:l_{i}\:]$}
			\STATE \texttt{UPDATE\_CS} ( $cl$ )
		\ENDIF
	\ENDFOR	
		
	\FOR {\textbf{each} label $l \in L$}
		\IF {$\mathbf{C}[\:l\:]$ is empty}
			\STATE $l_n \gets labels(V_t)[\:l\:]$
			\STATE $cl_{n}$ $\gets$ generate covering classifier with $V_t$ and $l_n$			
			\STATE insert $cl_{n}$ into the population $\mathbf{P}$
		\ELSIF{$(t - %\Bigl(
			\frac{\sum_{cl \in\mathbf{C}[\:l\:]}(cl.ts * cl.num)} %\big/ 
			{\sum_{cl \in \mathbf{C}[\:l\:]}cl.num} %\Bigr)
			> \theta_{GA}$)}
				\FOR{each classifier $cl$ in $\mathbf{C}[\:l\:]$}
					\STATE $cl.ts \leftarrow t $
				\ENDFOR
				\STATE $\{cl_{1},cl_{2}\} \leftarrow$ apply GA on $\mathbf{C}[\:l\:]$
				\STATE \texttt{ADD\_TO\_POPULATION} ( $cl_1$)
				\STATE \texttt{ADD\_TO\_POPULATION} ( $cl_2$)
			%\ENDIF
		\ENDIF
	\ENDFOR
	\WHILE {$\sum_{cl \in \mathbf{P}}cl.num > N $}
		\STATE delete rule from population $\mathbf{P}$ proportionally to $P_{del}$
	\ENDWHILE
\end{algorithmic}
\vspace{1mm}
\texttt{ADD\_TO\_POPULATION}~(~$cl$~)  
\begin{algorithmic}[1]
	\IF {$cl$ has non-zero coverage}
		\IF {$cl$ is not subsumed by parents}
			\IF {$cl$ is not subsumed by any rule in $\mathbf{P}$}
				\STATE insert $cl$ into the population $\mathbf{P}$
	 		\ENDIF
		\ENDIF
	\ENDIF
\end{algorithmic}
\end{algorithm}

\subsection{The Performance Component of \textsc{MlS-LCS} \label{sect:perf_comp}} 
Upon receipt of a data instance $V_t: X_t \rightarrow Y_t$, the system scans 
the current population of classifiers for those whose condition matches 
the input and forms the \emph{match set} $\mathbf{M}$. 
Next, for each label $l \in L$, a correct set $\mathbf{C}_l$ is formed 
containing the rules of $\mathbf{M}$ that correctly predict label $l$  
for the current instance\footnote{This is possible in a supervised framework, since the correct labels are directly available to the learning system.}. The classifiers in $\mathbf{M}$ incorrectly predicting label $l$ are placed in the 
 \emph{incorrect set} $\mathbf{!C}_l$. Finally, if the system is in test 
 mode\footnote{Under test mode, the population of \mlaslcs does not 
 undergo any changes; that is, the update and discovery components are 
 disabled.}, %and only the performance component is activated.}, 
a classification decision is produced based on the labels advocated by 
rules in $\mathbf{M}$ ($\mathbf{C}_l$ and $\mathbf{!C}_l$ cannot be 
produced since the actual labels are unknown).  

However, the process of classifying new samples based on models involving multi-label rules is not straightforward. In multi-label classification, a bipartition of relevant and irrelevant labels, rather than a single class, has to be decided upon, based on some  threshold. Furthermore, rulesets evolved with LCS may contain contradicting 
or low-accuracy rules. Therefore, a ``vote and threshold'' method is required to effectively classify unknown samples \citep{read2010scalable}. 
More specifically, an overall vote $w_l$ for each label $l$ $\in$ $L$ is obtained by allowing each rule to cast a positive (for advocated labels) or negative vote equal to its macro-fitness. Votes are cast only for labels that a rule provides concrete decisions for. 
The resulting votes vector $w$ is normalized, such that $\sum_{l \in L}\bar{w}_l=1$ and $\bar{w}_l \in [0,1], \forall l \in L$ and a threshold $t$ is used to select the labels that will be activated (those for which $\bar{w_l} \geq t$). Assuming that the thresholding method aims at activating at least one label, the range of reasonable thresholds is $t \in (0,0.5]$.

In our current work, we experimented with two threshold selection methods \citep{Yang2001, read2010scalable}, namely \emph{Internal Validation (\textsc{Ival})} and \emph{Proportional Cut (\textsc{Pcut})}.

\textbf{Internal Validation (\textsc{Ival})} selects, given the ruleset, the threshold value that maximizes a performance metric (such as accuracy), based on consecutive internal tests. It can produce good thresholds at a (usually) large computational cost, as the process of validating each threshold value against the training dataset is time-consuming. Its complexity, however, can be significantly improved by exploiting the fact that most metric functions are convex with respect to the threshold.  
 
 \textbf{Proportional Cut (\textsc{Pcut})} selects the threshold value that minimizes the difference in label cardinality $LCA$ (i.e., the mean number of labels that are activated per sample) between the training data and any other given dataset. This is achieved by minimizing the following error with respect to $t$: 
\begin{equation*} 
 	\text{\textbf{err} (t, LCA)} =\left| LCA(D) - 
	\left(\frac{1}{|G|}\sum_{i=1}^{|G|}\left|f_{th}(\bar{w_i},t)\right|\right)\right| 
\end{equation*} 
where $D$ is the training dataset, $f_{th}()$ is the threshold function and $G$ is the dataset with respect to which we tune the threshold $t$. It is worth noting that, in our case, it always holds that $G$$\subseteq$$D$. Tuning the threshold with respect to the test dataset would imply an a priori knowledge of label structure in unlabeled samples and would, thus, result in biased evaluations and, possibly, a wrong choice of models to be used for post-training predictions.
The \textsc{Pcut} method, although not tailored to any particular evaluation measure,  calibrates thresholds as effectively as \textsc{Ival}, at a fraction of the computational cost and is, thus, considered a method suitable for general use in experimental evaluations \citep{read2010scalable}.

Employing each rule's fitness as its confidence level, it is possible to predict the labels of new (unknown) data samples by using only the fittest rule of those matching each sample's attribute values. Of course, in case the fittest rule ``does not care'' for some of the labels, additional rules (sequentially, from a list of matching rules sorted by fitness) can be employed to provide a complete decision vector with specific values for all possible labels. The above described strategy, named \textbf{Best Rule Selection (\textsc{Best})}, has also been included in our experiments, since it is the one yielding the most compact, in terms of number of rules, prediction models.

%---------------------------- Update Component --------------------------------------------------------------------- 
\subsection{The Update Component of \textsc{MlS-LCS}}  \label{sec:update}
In training or explore mode, each classification of a data instance is 
associated with an update of the matching classifiers' parameters. 
More specifically: 
\begin{itemize}
\item 
[(i)] for all classifiers in match set $\mathbf{M}$, their experience 
	$exp$ is increased by one and their $msa$ value is updated, based on whether they provide a concrete decision; 
\item 
[(ii)] for all classifiers belonging to at least one correct set 
	$\mathbf{C}_l$, their correct set size $cs$ is updated, so that it 
	estimates the average size of all correct sets the classifier has 
	participated in so far; and 
\item 
[(iii)] all classifiers in match set $\mathbf{M}$ have their fitness $F$ updated. 
\end{itemize}
The specific update strategies for $fitness$  and correct set size 
$cs$ are presented in Alg. \ref{alg:updateFitness}.
%, wherein they are implemented in the \texttt{UPDATE\_FITNESS()} and \texttt{UPDATE\_CS()}  methods.   

\begin{algorithm}[htb]
\caption{Rule $fitness$ and $cs$ update for \mlaslcs}
\label{alg:updateFitness}
\texttt{UPDATE\_FITNESS} ( $cl$ ) 
\begin{algorithmic}[1]
  \FOR{\textbf{each} label $ l \in L$}
    \STATE $cl.tp \gets cl.tp + correctness(cl, l)$
    \STATE $cl.exp \gets cl.exp + 1$
    \STATE $cl.msa \gets cl.msa + msaValue(cl, l)$
  \ENDFOR
  \STATE $cl.acc \gets cl.tp/cl.msa$ \label{line:acc}
  \STATE $cl.F = (cl.acc)^\nu $ \label{line:k}
\end{algorithmic}
\vspace{1mm}
\texttt{UPDATE\_CS} ( $cl$ ) 
\begin{algorithmic}[1]
  \STATE $cs_{min} \gets {
  	\operatorname{min} \{\underset{cl \in \mathbf{C}_l}{\sum}{cl.num} \mid l \in L}\} $
  \STATE $cl.cs \gets cl.cs + \beta \left( cs_{min} - cl.cs\right)$
\end{algorithmic}
\end{algorithm}

\emph{Fitness calculation} in \mlaslcs is based on a supervised approach that involves computing the accuracy ($acc$) of classifiers as the percentage of their correct classifications (line~\ref{line:acc} of Alg.~\ref{alg:updateFitness}).  
Moreover, motivated by the need to distinguish between rules that 
provide concrete decisions (positive or negative) about labels and those whose decisions are ``indifferent'', we introduce the notion of $correctness$. 
The correctness value of a rule $cl$ for a label $l$ (with respect to a specific training cycle and, thus, specific $\mathbf{C}[\:l\:]$ and $\mathbf{!C}[\:l\:]$ sets) is calculated according to the following equation: 
\[ correctness(cl, l) =  \left\{ \begin{array}{l l}
1 & \text{if $cl \in \mathbf{C}[\:l\:]$} \\
0 & \text{if $cl \in \mathbf{!C}[\:l\:]$} \\
\omega & \text{if $cl \in (\mathbf{M} - \mathbf{C}[\:l\:] - \mathbf{!C}[\:l\:]$})\\
\end{array} \right.
\]
where $0 \le \omega \le 1$ for rules not deciding on $l$ for the current instance (i.e., for matching rules neither in $\mathbf{C}[\:l\:]$ nor in $\mathbf{!C}[\:l\:]$).

Accordingly, the match set appearances ($msa$) that a rule $cl$ obtains for a label $l$, during a specific training cycle, is differentiated depending on whether $cl$ provides a concrete decision or not, according to Eq.~\ref{eq:msa}, where $0 \le \phi \le 1$. % for rules not deciding on $l$ for the current instance.
\begin{equation} \label{eq:msa}
msaValue(cl, l) =  \left\{ \begin{array}{l l}
\phi & \text{if $cl \in ($$\mathbf{M}$-$\mathbf{C}[\:l\:]$-$\mathbf{!C}[\:l\:]$}) \\
1 & \text{otherwise} \\
\end{array} \right.
\end{equation}

In our current work, we explore a version of \mlaslcs that slightly penalizes ``indifferent'' rules by considering \#'s as partial ($\omega$=$0.9$) matches ($\phi$=$1$). The reasons that lead us to choose these specific values are detailed in Section \ref{sec:exp_art}. 
For now, though, let us again consider the simple example of the \toy problem and the rules of Eq.~\ref{eq:rules}. Supposing that both rules have not encountered any instances so far ($tp_1$=$tp_2$=$0$), when the system processes the instance $11000 \rightarrow 0100$, the rules' $tp$ values will become 1 and 0.9, respectively. This means that $r_1$'s fitness will be greater than that of $r_2$'s  when they compete in the GA selection phase for the first label and, thus, the system will have successfully applied the desired pressure towards maximally specific consequents.

Finally, as far as the update of \emph{rule overall correct-set size} is concerned, %(\texttt{updateCs(.)} method), 
%it is worth noting that 
we have chosen a rather strict estimation, employing the size of the smallest label correct set that the rule participates in. This choice is motivated by the need to exert fitness pressure in the population towards complete label-space coverage. This is, in our case, achieved by rewarding rules that explicitly advocate for or against ``unpopular'' labels.

%---------------------------- Discovery Component -------------------------------------------------------------------
\subsection{The Discovery Component of \textsc{MlS-LCS}}  \label{sec:disc_comp}
\mlaslcs employs two rule discovery mechanisms: a \emph{covering operator} and a \emph{steady-state niche genetic algorithm}. 

The \textbf{covering operator} is adapted from the one introduced in XCS \citep{XCS} and later used in UCS \citep{bernado2003, orriols08b} and most of their derivatives. It is activated only during training and introduces new rules to the population when the system encounters an empty correct set $\mathbf{C}[\:l\:]$ for a label $l$. Covering produces a single random rule with a condition matching the current input instance's attribute values and generalized with a given probability $P_\#$ per attribute. While this process is identical to the one employed in single-class LCS, it is followed by an additional generalization process applied to the rule consequent, which is essential to evolving generalized multi-label rules. All labels in the newly created rule's consequent are set to 0 or 1 according to the current input and  then generalized (converted to \#) with probability  $P_{label\#}$ per label, except for the current label $l$ that remains specific. 

The \textbf{genetic algorithm} is applied iteratively on all correct sets $ \mathbf{C}[\:l\:]$ and invoked at a rate $\theta_{GA}$, where $\theta_{GA}$ is defined as a (minimum) threshold on the average time since the last GA 
invocation of classifiers in $ \mathbf{C}[\:l\:]$ \citep{bernado2003}. The evolutionary process employs experience-discounted fitness-proportionate parent selection, with the selection probability $cl.P_{sel}$ assigned to each classifier $cl \in  \mathbf{C}[\:l\:]$ being calculated according to:
\begin{equation}
\label{eq:rouletteWheelSelection}
cl.P_{sel} = \frac{cl.num \cdot cl.F_{d}}{\underset{cl \in \mathbf{C}[\:l\:]}\sum{cl.num \cdot cl.F_{d}}}
\end{equation}
where
\begin{equation}
\label{eq:fitnessDiscount}
cl.F_{d} = \left\{
\begin{array}{ c l }
\displaystyle 0, & cl.exp < \theta_{exp}
\\
\displaystyle (cl.acc)^{\nu}, 	& \text{otherwise}
\end{array}
\right.
\end{equation}
and $\theta_{exp}$ is the experience threshold for fitness discounting.
After their selection, the two parent classifiers are copied to form two offspring, on which the \emph{multi-label crossover operator} and a uniform mutation operator are applied with probabilities $\chi$ and $\mu$, respectively.
%$\theta_{GA}$, $r$, $\chi$ and $\mu$ are set by the user.

%---------------------------- Multi-point Crossover ----------------------------------------------------------------
The \emph{multi-label crossover operator} is introduced in this work and intended for use specifically in multi-label classification settings. 
Its design was motivated by the fact that for the majority of datasets employed in our current work, the number of attributes is significantly larger than the number of labels (by at least one order of magnitude). This means that, using a single-point crossover operator, the probability that the crossover point would end up in the attribute space is significantly greater than that of it residing in the label space. Therefore, there would be a significantly greater probability of transferring the \textit{whole} consequent part from the parents to their corresponding offspring than that of transferring the decisions for only a subset of labels $L_{\chi} \subset L$. 

Actually, allowing the transfer of any \emph{set of decisions} as a policy for any given crossover occurring on $\mathbf{C}[\:l\:]$ would be a questionable choice: the fact that any two rules, selected to be parents, coincide in $\mathbf{C}[\:l_x\:]$ does not necessarily mean that they would coincide in $\mathbf{C}[\:l_y\:]$ where $x \neq y$ and $l_x, l_y \in L$. %except, of course, in the case where there is a-priori knowledge about it and it could be leveraged in order to increase the efficiency of the GA. 
Keeping this observation in mind, we designed the multi-label crossover operator, with the aim of exerting more pressure towards accurate decisions per label. The newly proposed operator achieves that \emph{by not transferring decisions from the selected parents to their corresponding offspring other than that about the current label}, i.e., the label corresponding to the correct set from which the parents were selected. %So, in essence, the multi-point crossover operator we introduce betters the LCS's performance by reducing the error in the rules' decision making.

More specifically, the crossover point is selected pseudo-randomly from the range $[0, cl.size_{\chi}$], where:
\begin{equation}
cl.size_{\chi} = cl.size - 2 \cdot (\abs{L} - 1)
\end{equation}
and $cl.size$ is the classifier size in bits.
This means that the multi-label crossover operator takes into account the rule's condition part and only one (instead of all $\abs{L}$) of its labels: the label for which the current correct set (on which the GA is applied) has been formed for. 
If the crossover point happens to be in the range $[0, cl.size_{\chi}-2$], that is in the condition part of the rule's chromosome, the two parent classifiers swap (a)~their condition parts beyond the crossover point and (b)~their decision for the current label from their consequent parts. Otherwise, that is when the crossover point happens to correspond to (any of the two bits representing) the current label, the two parent classifiers only swap their decision for the label being considered and no part of their conditions.

%---------------------------- Zero Coverage Prevention ------------------------------------------------------------- 
Returning to the GA-based rule generation process, after the crossover and mutation operators have been applied, \mlaslcs checks every offspring as per its ability to codify a part of the problem at hand. Given the supervised setting of multi-label classification, this is equivalent to checking that each rule covers at least one instance of the training dataset. The presence of rules in the population that fail to cover at least one instance, termed \emph{zero-coverage rules}\footnote{\emph{Coverage} is defined as the number of data instances a rule matches.}, is unnecessary to the system. Also, depending on the completeness degree of the problem, it may be hindering its performance by lengthening the training time and rendering the production rate of zero-coverage rules through the GA uncontrollable. Therefore, to avoid these problems, \mlaslcs removes zero-coverage rules just after their creation by the discovery component, assuring that $cl.coverage > 0 : \forall cl \in \mathbf{P}$.

Even after this step, the non-zero-coverage offspring are not directly inserted into the classifier population. They are gathered into a pool, until the GA has been applied to all label correct sets. Once the rule generation process has been completed for the current training cycle, and before their insertion into the classifier population, all rules in the offspring pool are checked for \emph{subsumption} (a) against each of their parents and (b) in case no parent subsumes them, against the whole rule population. If a classifier (parent or not) is found to subsume the offspring being checked, the latter is not introduced into the population, but the numerosity $num$ of the subsuming classifier is incremented by one instead. 
Subsumption conditions require that the subsuming classifier is sufficiently experienced ($cl.exp > \theta_{exp}$), accurate ($cl.acc > acc_0$) and \emph{more general} than the offspring being checked (with $\theta_{exp}$ and $acc_0$ being user-defined parameters of the system). Additionally, the \emph{generality condition} is extended for the multi-label case, such that a classifier $cl_i$ can only subsume a classifier $cl_j$, if $cl_i$'s condition part is equally or more general and its consequent part is equally or more specific than those, respectively, of the classifier $cl_j$ being subsumed.

%---------------------------- Coverage augmentation -----------------------------------------------------------------
\subsection{Population control strategies employed in \textsc{\mlaslcs}}
\label{sec:pop_control}
The system maintains an upper bound on the population size (at the microclassifier level) by employing a \textbf{deletion mechanism}, according to which a rule $cl$ is selected for deletion with probability $P_{del}$:
\begin{equation} \label{eq:U2}
cl.P_{del} = \frac{cl.num \cdot cl.d}{\sum\limits_{cl_i\in \mathbf{P}}cl_i.num \cdot cl_i.d}
\end{equation}
where
\begin{equation*} \label{eq:U3}
cl.d = \left\{
\begin{array}{ c l }
\displaystyle e^{(cl.F)^{-1}}, & \text{if } cl.exp < \theta_{del} \\
\displaystyle (e^{(cl.cs-1)})/{cl.F}, & \text{otherwise} 
\end{array}
\right.
\end{equation*}
and $\theta_{del}$ is a user-defined experience threshold.

In addition to the deletion mechanism that is present in most LCS implementations, in \mlaslcs we introduce a new \textbf{population control strategy} that aims to increase the mean coverage of instances by the rules in the population. This strategy corresponds to the ``control match set $\mathbf{M}$'' step (line~\ref{func:controlMatchSet} of Alg.~\ref{alg:mlaslcs}) in the overall training cycle of \mlaslcs and is based on the following observations:
\begin{itemize}
\item Given a set of rules, such as the match set $\mathbf{M}$, the rules it comprises lie on different coverage levels. This means that rules cover different numbers of dataset instances, depending on the degree of generalization that the LCS has achieved.
\item A given coverage level $cov\_level_i$ in $\mathbf{M}$ (a subset of rules in $\mathbf{M}$ whose members cover the same number of instances) holds rules of various fitnesses.
\item If there are two or more rules in the lowest coverage level $cov\_level_{min}$ in $\mathbf{M}$,
the rule $cl_j$ whose fitness is the lowest among them is not necessary in $\mathbf{M}$. That is because there exist more rules that cover the instance from which $\mathbf{M}$ was generated and are, 
in addition,  more fit overall, classifying instances more accurately. The rule may still be of use in $\mathbf{P}$, if it is the sole rule covering an instance in the population. However, in the general case, $cl_j$ can be removed from the population $\mathbf{P}$ without any considerable loss of accuracy for the system.
\end{itemize}

The invocation condition for the \emph{match set control strategy} (line~\ref{condition:rouletteWheelDeletionsCommenced} of Alg.~\ref{alg:mlaslcs}) means that the corresponding deletion mechanism will only be activated after the population
has reached its upper numerosity boundary for the first time. Thus, ``regular'' deletions from the population and deletions of low-coverage rules from the match set are two processes (typically) applied simultaneously in the system. 
Using the above invocation condition accomplishes two objectives: (i)~it prevents, during the first training iterations, the deletion of fit and specific rules that could be pass their `useful' genes on to the next generation and (ii)~it prevents (to a certain degree) the deletion of rules that coexist with others in the lowest coverage level of a specific match set but are unique in another. 

Finally, as far as the computational cost of implementing population control is concerned, it is worth noting that it is negligible, as the coverage value for each rule is gradually determined through a full pass of the training dataset and is used only after that point (from when on, it does not change).

\subsection{Clustering-based initialization of \mlaslcs}
\label{sec:init}
 
\mlaslcs also employs a population initialization method that extracts information about the structure of the studied problems through a pre-training clustering phase and exploits this information by transforming it into rules suitable for the initialization of the learning process. The employed method is a generalization for the multi-label case of the clustering-based initialization process presented in \citet{tzima2012} that has been shown to boost LCS performance, both in terms of predictive accuracy and the final evolved ruleset’s size, in supervised single-label classification problems. 

Simply put, the \emph{clustering-based initialization method} of \mlaslcs detects a representative set of ``data points'', termed centroids, from the target multi-label dataset $\textbf{D}$ and transforms them into rules for the initialization of the LCS rule population prior to training. 

More specifically,

the dataset $\textbf{D}$ is partitioned into $N$ subsets, 
where $N$ is the total number of discrete label combinations present in $\textbf{D}$. 
Each subset $Partition_i$ consists of the instances whose label combination
matches the discrete label combination $i$. %, such that:
For each partition $Partition_i$, $1$$\leq$$i$$\leq$$N$:
\begin{itemize}
	\item
	The instances belonging to the partition are grouped into $M_{i} = \left\lceil\gamma \cdot \abs{Partition_{i}}\right\rceil$ clusters, 
	where $\abs{Partition_{i}}$ is the number of instances in the $i^{th}$ partition and $\gamma$ ($0<\gamma\leq 1$) is a user-defined parameter.
	\item
	For each cluster $cluster_{j}$, $1\leq j \leq M_{i}$, identified in the previous step, its centroid is found employing a clustering algorithm (in our case, the k-means algorithm).
	Then, a new rule is created whose condition part matches the centroid's attribute values (more details on this procedure can be found in \citet{tzima2012}), while the decision part is set to the discrete label combination associated with the current partition. The centroid-to-rule transformation process also includes a generalization step (similar to the one used by the covering operator): some of the newly created rule's conditions and decisions are generalized (converted to ``don't cares''), taking into account the attribute $P_{\# init}$ and label $P_{label\# init}$ generalization probabilities defined by the user for clustering.
\end{itemize} 
Finally, all $K = \sum\limits_{i=1}^N M_{i}$ rules %of the form
%$$rule_{ij}: condition_{ij} \rightarrow distinct\_label\_combination_{i}$$ 
created by clustering the training dataset are merged to
create the ruleset used to initialize the learning process.

In our current work, we chose not to experiment with tuning the clustering-based initialization process parameters and used the following values for all reported experiments: $\gamma=0.2$, $P_{\# init}=0$ and $P_{label\# init}=0$.

\section{Experimental Validation of \mlaslcs}
\label{sec:evaluation}
In this Section, we present an experimental evaluation of our proposed multi-label LCS approach\footnote{The Java source code of our implementation of \mlaslcs used throughout all reported experiments is publicly available at: https://github.com/fanioula/mlslcs.}. We first provide a brief analysis of \mlaslcs's behavior on two artificial datasets and then compare its performance to that of 6 other state-of-the-art methods on 7 real-world datasets.

\subsection{Experiments on artificial datasets}\label{sec:exp_art}
Since the focus of our current work is on multi-label classification, we begin our analysis with two artificial problems, named \toy and \pos respectively, that we consider representative of a wide class of problems from our target real-world domain, in terms of the label correlations involved. The \toy problem has already been described in Section \ref{sec:genMLrules}. The \posN (here, N=4) problem has $N$ binary attributes and $N$ labels. In every data sample, only one label is active, that is the label $k$ corresponding to the most significant bit of the binary number formed by the sample's attributes. It is evident that, in this case, there is great imbalance among the labels, since label $l_1$ is only activated once, while label $l_{N}$ is activated in $2^{N-1}$ instances. The shortest complete solution 
of the problem involves exactly $N$+$1$ rules, with different degrees of generalization in their condition parts, but no generalizations in their consequent parts. Specifically, for the \pos problem the shortest complete solution (SCS) includes the following rules: 
\begin{equation*}
\begin{split}
0000 	& \rightarrow  0000  \\
0001 	& \rightarrow 0001 \\
001\# 	& \rightarrow 0010 \\ 
01\#\# 	& \rightarrow 0100 \\
1\#\#\# 	&\rightarrow 1000
\end{split}
\end{equation*}

Overall, one can easily observe that \toy is a problem where two of the labels are only dependent on attribute values and independent of other labels, while \pos involves labels that are completely dependent on each other (in fact, they are mutually exclusive). Most (non-trivial) real-world problems will be a ``mixture'' of these two cases (i.e., will involve a mixture of uncorrelated and correlated labels), so our intention is to tune the system to perform as well as possible for both artificial problems. In the current paper, we focus our study on the fitness update process (see Section \ref{sec:update}) and, more specifically, the choice of the $\omega$ parameter value, given that we consider ``don't cares'' as partial matches ($\omega$$\le$$1$,  $\phi$=$1$). 

Regarding performance metrics, the percentage of the SCS was selected as an appropriate performance metric, indicative of the progress of genetic search. Along with the \scs, we also report the multi-label accuracy (Eq.~\ref{eq:acc}) achieved by the system throughout learning and the average number of rules in the final models evolved. All reported results are averaged over 30 runs (per problem and $\omega$ parameter setting) with different random seeds.

For all experiments and both problems, we kept the majority of parameter settings fixed, using a typical setup,  consistent with those reported in the literature of single-label LCS: $\mu$=$0.04$, $\chi$=$0.8$, $\beta$=$0.2$, $\nu$=$10$, $k$=$5$, $\theta_{del}$=$20$, $\theta_{exp}$=$10$, and $acc_0$=$0.99$. The population size $|P|$ was set to $10^4$, the number of iterations $I$ was $1500*64$, the GA invocation rate $\theta_{GA}$ was $2000$ and the generalization probability $P_{\#}$ was $0.33$. The only parameter varied between the two problems was the label generalization probability $P_{label\#}$ which was set to $0.5$ and $0.2$, respectively, for \toy and \pos. 

\begin{figure}[tb]
\centering
\subfigure[\scs achieved by \mlaslcs in \toy ]{
\includegraphics[trim=0cm 0.6cm 0cm 0.5cm, clip=true, width=0.95\columnwidth]{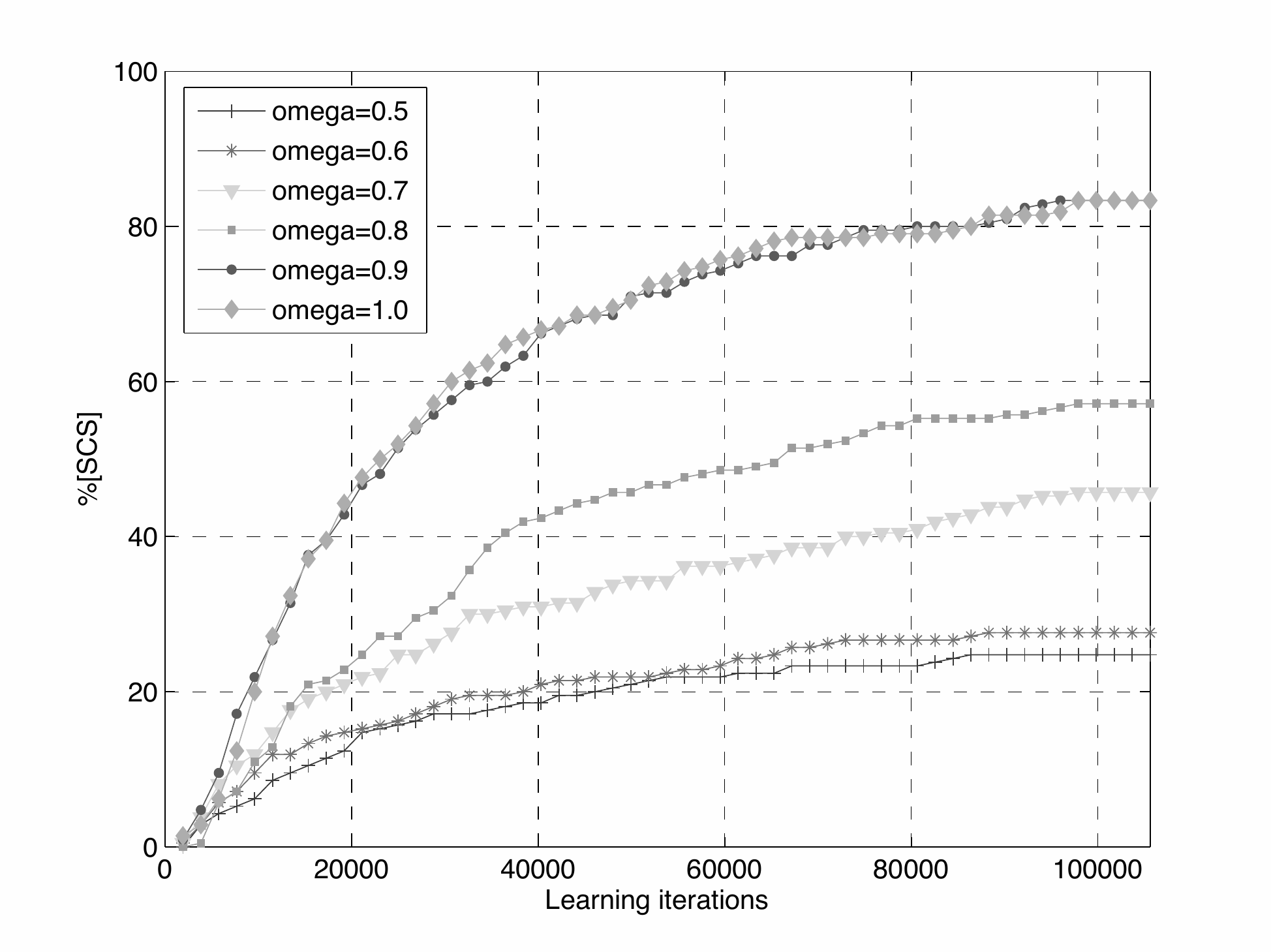}
\label{fig:vBAM}
}
\\
\subfigure[Accuracy achieved by \mlaslcs in \toy ]{
\includegraphics[trim=0cm 0.6cm 0cm 0.5cm, clip=true, width=0.95\columnwidth]{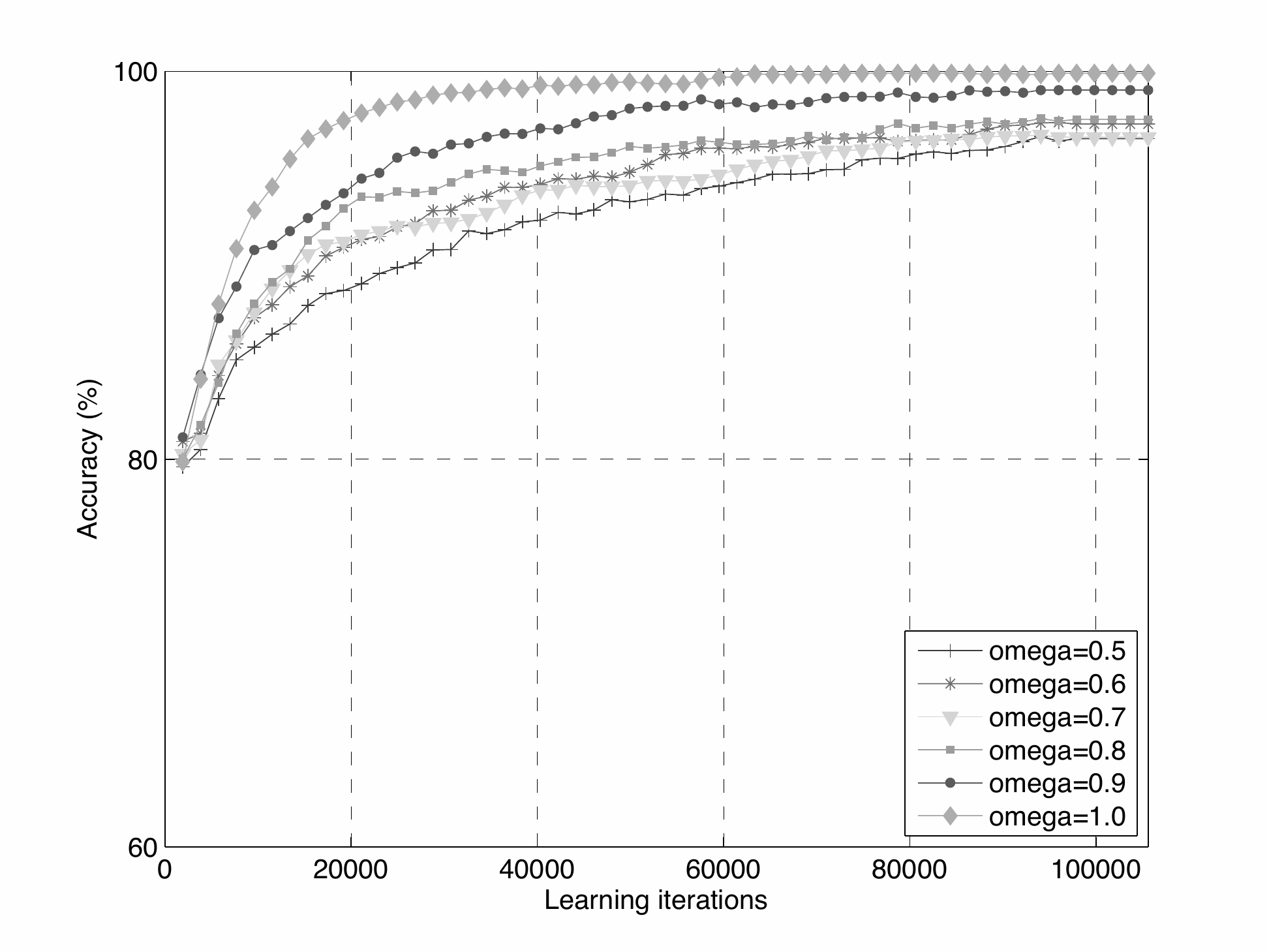}
\label{fig:vAcc}
}
\caption{Percentage of the shortest complete solution (\scs) and multi-label accuracy achieved throughout the learning process for the\toy problem. All curves are averages over thirty runs. \label{fig:allCurves}}
\end{figure}

\begin{figure}[tb]
\centering
\subfigure[\scs achieved by \mlaslcs in \pos ]{
\includegraphics[trim=0cm 0.6cm 0cm 0.5cm, clip=true, width=0.95\columnwidth]{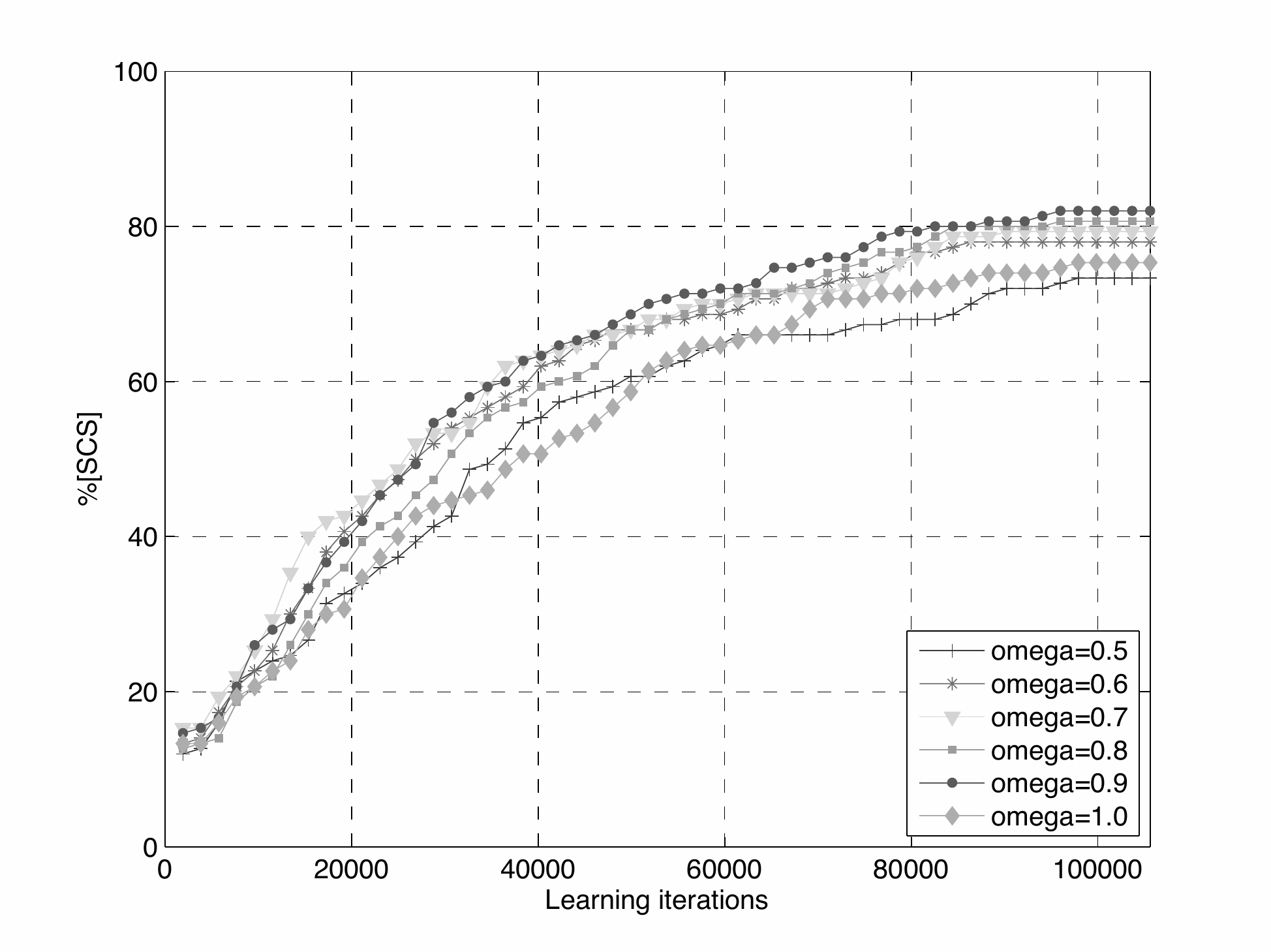}
\label{fig:vBAM}
}
\\
\subfigure[Accuracy achieved by \mlaslcs in \pos ]{
\includegraphics[trim=0cm 0.6cm 0cm 0.5cm, clip=true, width=0.95\columnwidth]{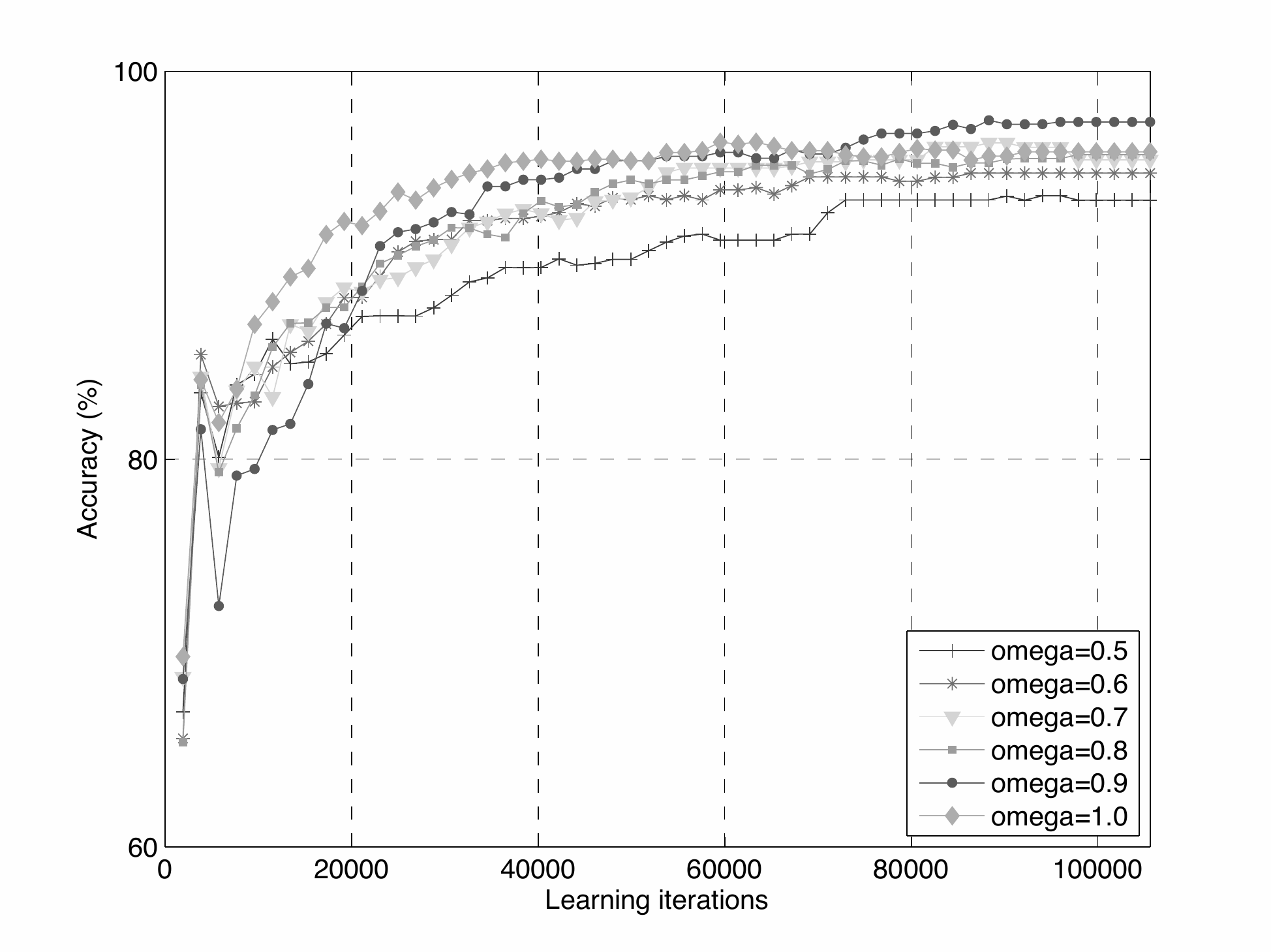}
\label{fig:vAcc}
}
\caption{Percentage of the shortest complete solution (\scs) and multi-label accuracy achieved throughout the learning process for the \pos problem. All curves are averages over thirty runs. \label{fig:allCurves2}}
\end{figure}

The results of our experiments, depicted in Figures \ref{fig:allCurves} and \ref{fig:allCurves2}, reveal that the value of the $\omega$ parameter affects both the accuracy and the quality (in terms of number of rules) of the evolved solutions. This is especially evident in the \toy problem, where the SCS contains a complex trade-off of feature-space generality and label-space specificity: low values of $\omega$ result in over-penalizing label-space indifferences and exerting pressure for highly specific consequents, thus also adversely affecting the system's accuracy. The same pressure towards consequent specificity proves beneficial in the \pos problem, due to the nature of the problem's SCS that comprises rules providing specific decisions for all labels. 

Given our goal to optimize system performance for both problems and the importance of the accuracy metric in real-world applications, choosing the value $\omega$=$0.9$ is a good trade-off. For this value and the \toy problem, the system discovers $83.33$\% of the SCS on average  ($100$\% in $14$ out of the $30$ averaged experiments) and achieves a $99.03$\% accuracy ($100$\% in $28$ of the averaged experiments). For the \pos problem (and again $\omega$=$0.9$), the system discovers $82$\% of the SCS on average ($100$\% in $10$ experiments) and achieves a $97.38$\% accuracy ($100$\% in $25$ experiments). 

As far as the size of the final rulesets is concerned (again for $\omega$=$0.9$), after applying a simple ruleset compaction strategy (i.e., ordering the rules by their macro-fitness and keeping only the top rules necessary to fully cover the training dataset and have specific decisions for all labels), we get models with $34.53$ and $9.87$ rules on average for \toy and \pos, respectively. This points to aspects of the rule evaluation process that need to be further investigated, since for the chosen value of $\omega$, some of the SCS rules are present, but not prevalent enough in the final rulesets for the compacted solutions to be of the optimal size.

\subsection{Experimental setup for real-world problems}
The benchmark datasets employed in this set of experiments are listed in Table~\ref{tab:datasetStats}, along with their associated descriptive statistics and application domain. %and references (where applicable). 
The datasets are ordered by complexity ($|D| \times |L| \times |X|$), while Label Cardinality (LCA) is the average number of labels relevant to each instance. 
%The Proportion of Dist inct label combinations (PDist) is simply the number of distinct label subsets relative to the total number of examples: 
We strived to include a considerable variety and scale of multi-label datasets. In total, we used 7 datasets, with dimensions ranging from 6 to 174 labels, and
from less than 200 to almost 44,000 examples. All of the datasets are readily available from the Mulan web-site (http://mulan.sourceforge.net/datasets.html).

\begin{table*}[htbp]
\centering 
\caption{Benchmark datasets, along with their application domain and statistics: number of examples $|D|$, number of nominal (c) or numeric (n) attributes $|X|$, number of labels $|L|$, number of distinct label combinations \textsc{DIST}, label density \textsc{DENS} and cardinality \textsc{LCA}. Datasets are ordered by complexity,  defined as $|D| \times |L| \times |X|$. \label{tab:datasetStats}}
%\begin{tabular}{lrrrrrrlcr}
\begin{tabular}{lrrrrrrlr}
\toprule
dataset	& \multicolumn{1}{c}{$|D|$}	& \multicolumn{1}{c}{$|X|$}	& $|L|$ 	&  \textsc{DIST}	& \textsc{DENS}		
& \textsc{LCA} & domain %& references 
& complexity
\\
\midrule
flags 	& 194		& 9C+10N	&7			&54 			& 0.485		
& 3.39	& images		%& \citep{goncalves13}			
& 2.58E+04
\\
emotions	& 593		& 72N		& 6			& 27			& 0.311		
& 1.87	& music		%& \citep{trohidis2008multilabel} 	
& 2.56E+05
\\
genbase	& 662		& 1186C		& 27			& 32			& 0.046		
& 1.25	& biology		%& \citep{diplaris2005protein}		
& 2.00E+06
\\
%yeast	& 2417		& 103N		& 14			& 198		& 0.303		& 4.24	& biology		%& \citep{elisseeff2002kernel} & 3.49E+06\\
scene	& 2407		& 294N		& 6			& 14			& 0.179		
& 1.07	& images		%& \citep{boutell2004}			
& 4.25E+06
\\
%medical	& 978		& 1449 C		& 45			& 94			& 0.03		& 1.25\\
CAL500 	& 502		& 68N		& 174		& 502		& 0.150		
& 26.04	& music		%& \citep{turnbull2008semantic} 	
& 5.94E+06
\\
enron%\footnote{http://bailando.sims.berkeley.edu/enron\_email.html}	
		& 1702		& 1001C		& 53			& 753		& 0.064		
		& 3.38	& text		%& 							
		& 9.03E+07
		\\
mediamill 	& 43907	& 120N		&101		& 6555 		& 0.043		
& 4.38 	&video		%& \citep{snoek2006}			
& 5.32E+08 
\\
\bottomrule
\end{tabular}
\end{table*}

Evaluation is done in the form of ten-fold cross validation for the four smallest datasets\footnote{The specific splits in folds, along with the detailed results of the rival algorithm parameter tuning phase, are available at http://issel.ee.auth.gr/software-algorithms/mlslcs/.}. For the \emph{enron}, \emph{CAL500} and \emph{mediamill} datasets a train/test split (provided on the Mulan website) is used instead, since cross-validation is too time and/or computationally intensive for some methods\footnote{Some of the rival algorithms’ runs could not be completed, even on a machine with 64GB of RAM.}. 

The rival algorithms against which the proposed \mlaslcs algorithm is compared are HOMER, \rakel, ECC, CC, \mlknn and BR-J48. For all algorithms, except ECC and CC, their implementations provided by the Mulan Library for Multi-label Learning \citep{tsoumakas2010mulan} were used, while for ECC and CC we used the MEKA environment (http://meka.sourceforge.net/). 

As far as the parameter setup of the algorithms is concerned, in general, we followed the recommendations from the literature, combined with a modest parameter tuning phase, where appropriate. More specifically:
\begin{itemize}
\item BR refers to a simple binary-relation transformation  of each problem using the C4.5 algorithm (WEKA's \citep{weka} J48 implementation) and serves as our baseline. 
\item For HOMER, Support Vector Machines (SVMs) are used as the internal classifier (WEKA’s SMO implementation). For the number of clusters, five different values (2-6) are considered and the best result is reported. 
\item We experiment with three versions of \rakel and report the best result: (a) the default setup (subset size $k=3$ and $m=2L$ models) with C4.5 (WEKA's J48) as the baseline classifier, (b) the ``extended setup'', with a subset size equal to half the number of labels $k=|L|/2$ and $m=min(2|L|, 100)$ models, and C4.5 (WEKA's J48 implementation) as the baseline classifier, and (c) the ``extended setup'' and SVMs (WEKA's SMO implementation) as the baseline classifier. 
\item ECC and CC are used with SVMs (WEKA's SMO implementation) as the baseline classifier, while the number of models for ECC is set to 10, as proposed by the algorithm's authors in \citep{read2009classifier}. 
\item Finally, the number of neighbors for the \mlknn method is determined by testing the values 6 through 20 (with step 2) and selecting the best result per dataset. 
\end{itemize}
 %Overall, it is also important to note that, 
 Where not stated differently, the default parameters were used.

For \mlaslcs, we kept the majority of parameters fixed through all experiments, using the typical setup reported for the artificial problem experiments. The parameters varied were 
the population size $|P|$, the number of iterations $I$, the GA invocation rate $\theta_{GA}$ and the generalization probabilities $P_{\#}$ and $P_{label\#}$. The choice of specific parameter values 
(Table~\ref{tab:paramValues}) was based on an iterative process that involved starting with default values for all parameters ($|P|$=$5000$, and $I$=$500$*$|D|$, $\theta_{GA}$=$2000$, $P_{label\#}$=$0.1$,  $P_{\#}$=$0.5$) and tuning one parameter at a time, according to the following steps: 
\begin{enumerate}
\item %the label generalization probability 
$P_{label\#}$ was set to either 0.1 or 0.01, depending on the resulting model's performance on the train dataset;
\item for %the attribute generalization probability 
$P_{\#}$ the values 0.33, 0.4, 0.8, 0.9, and 0.99 were iteratively tested and the one leading to the greater coverage of the train dataset's instances was selected;
\item %the GA invocation rate 
$\theta_{GA}$ was selected between the values 300 and 2000, based on which one of them leads to a faster suppression of the covering process;
\item the population size $|P|$ was selected among the values 1000, 2000, 9000, 12000, and 25000, based on the resulting model's performance on the train dataset;
\item %as far as the number of iterations $I$ is concerned, 
evolved models were evaluated every $I$=$100$*$|D|$ iterations and training stopped when the performance on the test dataset (with respect to the accuracy metric) was greater than that of the baseline BR approach. 
\end{enumerate}
During the tuning process, the parameter values selected in each step were used (and kept constant) in all subsequent steps.
\begin{table}[htbp]
\centering 
\caption{\mlaslcs parameters for the benchmark datasets
%: population size $|P|$, GA invocation rate $\theta_{GA}$ and generalization probabilities $P_{\#}$ and $P_{label\#}$. 
\label{tab:paramValues}}
\begin{tabular}{lrrrrrr}
\toprule
Dataset	&$I/|D|$ 	& $|P|$	& $\theta_{GA}$	 & $P_{\#}$	& $P_{label\#}$ \\
\midrule
flags		& 500 	&1000 	& 2000	& 0.33	& 0.01 \\
emotions	& 500 	&5000	& 2000	& 0.8	& 0.01 \\
genbase	& 500	&12000	& 2000	& 0.4	& 0.10 \\
%yeast	& 18000	& 4000	& 0.85	& 0.01 \\
scene	& 2500	&9000	& 300	& 0.99	& 0.10 \\
%medical	& 2500	& 2000	& 0.99	& 0.10 \\
CAL500 	& 200	&1000	& 2000	& 0.9	& 0.10 \\
enron	& 600 	&25000	& 2000	& 0.99	& 0.10 \\
mediamill &10 	&1000	& 2000	& 0.9	&0.10\\
\bottomrule
\end{tabular}
\end{table}
It is also worth noting that, when using \textsc{Ival} for 
\mlaslcs, the corresponding thresholds were calibrated based 
on the (multi-label) accuracy metric, as in \rakel.

Regarding the statistical significance of the measured differences in algorithm performance, we employ the procedure suggested in \citep{Demsar} for robustly comparing classifiers across multiple datasets. This procedure involves the use of the \emph{Friedman test} %\citep{Friedman1937}  %,Friedman1940}  
to establish the significance of the differences between classifier ranks and, potentially, a post-hoc test to compare classifiers to each other. In our case, where the goal is to compare the performance of all algorithms to each other, %ii) compare the two versions of the proposed \mlaslcs algorithm. For the first goal 
the \emph{Nemenyi test} %\citep{nemenyi63} 
was selected as the appropriate post-hoc test. 

\subsection{Comparative Analysis of Results}

Table \ref{tbl:resultsMLSLCS} summarizes the results for the \mlaslcs algorithm for all inference methods (see Section \ref{sect:perf_comp}), namely Proportional Cut (\textsc{Pcut}), Internal Validation (\textsc{Ival}) and Best Classifier Selection (\textsc{Best}), all three evaluation metrics (multi-label accuracy, exact match and Hamming loss) and all datasets used in this study. All values reported are at a \% scale and the results for the three evaluation metrics for each inference method refer to the same experiment per dataset.

\begin{table*}[htbp]
\centering 
\caption{Evaluation results for all inference methods employed by the \mlaslcs algorithm and all metrics used in algorithm comparisons. 
The best value per problem-metric pair is marked in bold. 
\label{tbl:resultsMLSLCS} }
\begin{tabular}{lccc|ccc|ccc}
\toprule
& \multicolumn{3}{c}{Accuracy} &  
\multicolumn{3}{c}{Exact Match} & \multicolumn{3}{c}{Hamming Loss} \\ 
& \textsc{Pcut} & \textsc{Ival} & \textsc{Best} & \textsc{Pcut} & \textsc{Ival} & \textsc{Best} & \textsc{Pcut} & \textsc{Ival} & \textsc{Best}\\
\midrule							
flags	&$63.46$	&$\mathbf{64.33}$	&$58.19$	&$\mathbf{22.90}$	&$21.81$	&$19.04$	&$\mathbf{23.93}$	&$24.22$	&$26.88$\\
emotions	&$59.34$	&$\mathbf{59.90}$	&$43.90$	&$\mathbf{35.28}$	&$34.97$	&$21.13$	&$\mathbf{18.97}$	&$19.08$	&$28.49$\\
genbase	&$98.46$	&$98.69$	&$\mathbf{98.91}$	&$96.99$	&$97.28$	&$\mathbf{97.88}$	&$00.13$	&$\mathbf{00.11}$	&$\mathbf{00.11}$\\
scene	&$\mathbf{66.61}$	&$66.49$	&$57.99$	&$57.58$	&$\mathbf{58.13}$	&$54.18$	&$\mathbf{20.99}$	&$21.85$	&$26.50$\\
CAL500	&$32.45$	&$33.24$	&$\mathbf{93.21}$	&$00.30$	&$00.30$	&$\mathbf{91.02}$	&$11.09$	&$\mathbf{10.93}$	&$14.10$\\
enron	&$38.92$	&$\mathbf{38.93}$	&$29.41$	&$06.04$	&$\mathbf{07.25}$	&$03.28$	&$15.40$	&$18.39$	&$\mathbf{01.91}$\\
mediamill	&$33.97$	&$\mathbf{34.70}$	&$21.16$	&$00.89$	&$\mathbf{01.97}$	&$00.59$	&$06.13$	&$\mathbf{05.73}$	&$07.46$\\
\bottomrule
\end{tabular}
\end{table*}

Inspecting the obtained results, one can easily conclude that while no inference method is clearly dominant, \textsc{Ival} seems to yield the best results overall. It is also worth noting that the \textsc{Best} method outperforms the other two inference methods for 2 out of the 7 studied datasets, although it involves a considerably smaller number of rules in its final models. Especially in the case of the CAL500 dataset, the use of the full evolved ruleset (thresholded through \textsc{Pcut} or \textsc{Ival})  seems to be particularly harmful for system performance. This indicates a problem with either the evolution of rules or the threshold selection procedures that needs to be further investigated in the future.

In general, results with the \textsc{Best} method are acceptable and close to that of the other inference methods. Thus, the considerably smaller rulesets involved in \textsc{Best} models can be considered an effective summary of the target problem's solution to be used for descriptive purposes. The need for such ``description'' is especially evident in real-world classification problems, where the desired solution must be interpretable by human experts and/or decision makers.

Considering the experiment that corresponds to the inference method with the best accuracy value for our proposed \mlaslcs algorithm, Tables \ref{tbl:experiments1} - \ref{tbl:experiments3} summarize the results of comparing it with its rival learning  techniques.  Achieved values (\%) for the three evaluation metrics (multi-label accuracy, exact match and Hamming loss) and all datasets used in this study are reported. 
In Table \ref{tbl:experiments1} along with the accuracy rates, we also report each algorithm's overall average rank (row labeled ``Av. Rank'') and its position in the final ranking (row labeled ``Final Pos.''). Accordingly, Tables \ref{tbl:experiments2} and \ref{tbl:experiments3}, respectively, report the values for the exact match and the Hamming loss metrics, along with the corresponding rankings.

\begin{table*}[htbp]
\centering 
\caption{Algorithm comparison based on multiple evaluation metrics. 
Superscripts refer to algorithm ranks 
(per problem) according to the Friedman test, the column labeled ``Av. Rank'' reports the 
average rank of the method in the corresponding row, while the column labeled ``Final Pos.'' 
holds its position in the (overall) final ranking.}

\subtable[Algorithm evaluation based on the ``Accuracy'' metric.]
{
\label{tbl:experiments1}
\begin{tabular}{lccccccc}
\toprule
& \multicolumn{1}{c}{HOMER} & 
\multicolumn{1}{c}{\rakel} & \multicolumn{1}{c}{ECC} & \multicolumn{1}{c}{CC} & \multicolumn{1}{c}{\mlknn} & \multicolumn{1}{c}{BR} &\multicolumn{1}{c}{\mlaslcs} \\ 
\midrule							
flags		&$63.64^{2.0}$		&$61.48^{5.0}$	&$59.13^{7.0}$	&$59.70^{6.0}$	&$61.91^{3.0}$	&$61.57^{4.0}$	&$64.33^{1.0}$\\
emotions	&$59.18^{3.0}$		&$58.74^{4.0}$	&$59.45^{2.0}$	&$56.18^{5.0}$	&$54.86^{6.0}$	&$44.33^{7.0}$	&$59.90^{1.0}$\\
genbase	&$99.07^{2.0}$		&$99.02^{3.0}$	&$98.58^{6.0}$	&$99.28^{1.0}$	&$97.99^{7.0}$	&$98.62^{5.0}$	&$98.91^{4.0}$\\
scene	&$65.58^{6.0}$		&$67.56^{3.0}$	&$70.72^{1.0}$	&$66.48^{5.0}$	&$68.23^{2.0}$	&$54.47^{7.0}$	&$66.61^{4.0}$\\
CAL500	&$35.40^{5.0}$		&$92.78^{2.0}$	&$34.20^{6.0}$	&$24.20^{7.0}$	&$36.40^{4.0}$	&$80.92^{3.0}$	&$93.21^{1.0}$\\
enron	&$41.23^{4.0}$		&$45.67^{1.0}$	&$44.80^{2.0}$	&$41.30^{3.0}$	&$35.30^{7.0}$	&$36.71^{6.0}$	&$38.93^{5.0}$\\
mediamill&$42.83^{2.0}$		&$44.98^{1.0}$	&$39.70^{4.0}$	&$37.30^{5.0}$	&$42.50^{3.0}$	&$36.89^{6.0}$	&$34.70^{7.0}$\\
\midrule
Av. Rank	&$3.43$		&$2.71$	&$4.00$	&$4.57$	&$4.57$	&$5.43$	&$3.29$\\
Final Pos.	&$3.0$		&$1.0$	&$4.0$	&$5.5$	&$5.5$	&$7.0$	&$2.0$\\
\bottomrule
\end{tabular}
}

\subtable[Algorithm evaluation based on the ``Exact Match'' metric.]
{
\label{tbl:experiments2}
\begin{tabular}{lccccccc}
\toprule
& \multicolumn{1}{c}{HOMER} &  
\multicolumn{1}{c}{\rakel} & \multicolumn{1}{c}{ECC} & \multicolumn{1}{c}{CC} & \multicolumn{1}{c}{\mlknn} & \multicolumn{1}{c}{BR} &\multicolumn{1}{c}{\mlaslcs} \\  
\midrule							
flags		&$18.65^{3.0}$		&$20.11^{2.0}$	&$16.20^{6.0}$	&$18.24^{4.0}$	&$12.90^{7.0}$	&$16.25^{5.0}$	&$21.81^{1.0}$\\
emotions	&$31.36^{5.0}$		&$34.07^{3.0}$	&$34.75^{2.0}$	&$32.27^{4.0}$	&$30.02^{6.0}$	&$16.70^{7.0}$	&$34.97^{1.0}$\\
genbase	&$98.18^{2.0}$		&$98.03^{3.0}$	&$97.13^{6.0}$	&$98.48^{1.0}$	&$96.53^{7.0}$	&$97.28^{5.0}$	&$97.88^{4.0}$\\
scene	&$59.70^{5.0}$		&$61.33^{4.0}$	&$61.89^{3.0}$	&$62.03^{2.0}$	&$63.19^{1.0}$	&$44.04^{7.0}$	&$57.58^{6.0}$\\
CAL500	&$00.00^{5.5}$		&$20.36^{2.0}$	&$00.00^{5.5}$	&$00.00^{5.5}$	&$00.00^{5.5}$	&$0.30^{3.0}$		&$91.02^{1.0}$\\
enron	&$12.26^{3.0}$		&$16.41^{1.0}$	&$11.20^{4.0}$	&$12.40^{2.0}$	&$08.98^{5.0}$	&$8.64^{6.0}$		&$07.25^{7.0}$\\
mediamill &$06.03^{5.0}$		&$13.13^{1.0}$	&$09.40^{3.0}$	&$08.70^{4.0}$	&$12.27^{2.0}$	&$5.35^{6.0}$		&$01.97^{7.0}$\\
\midrule
Av. Rank	&$4.07$		&$2.29$	&$4.21$	&$3.21$	&$4.79$	&$5.57$	&$3.86$\\
Final Pos.	&$4.0$		&$1.0$	&$5.0$	&$2.0$	&$6.0$	&$7.0$	&$3.0$\\
\bottomrule
\end{tabular}
}

\subtable[Algorithm evaluation based on the ``Hamming Loss'' metric.]
{
\label{tbl:experiments3}
\begin{tabular}{lccccccc}
\toprule
& \multicolumn{1}{c}{HOMER} & 
\multicolumn{1}{c}{\rakel} & \multicolumn{1}{c}{ECC} & \multicolumn{1}{c}{CC} & \multicolumn{1}{c}{\mlknn} & \multicolumn{1}{c}{BR} &\multicolumn{1}{c}{\mlaslcs} \\ 
\midrule							
flags		&$24.98^{2.0}$		&$25.28^{3.0}$	&$26.29^{7.0}$	&$25.91^{6.0}$	&$25.79^{5.0}$	&$25.59^{4.0}$	&$24.22^{1.0}$\\
emotions	&$19.24^{5.0}$		&$18.21^{1.0}$	&$18.48^{2.0}$	&$20.97^{6.0}$	&$18.99^{3.0}$	&$24.89^{7.0}$	&$19.08^{4.0}$\\
genbase	&$00.08^{2.0}$		&$00.07^{1.0}$	&$00.12^{6.0}$	&$00.08^{3.0}$	&$00.15^{7.0}$	&$00.11^{4.0}$	&$00.11^{5.0}$\\
scene	&$11.45^{6.0}$		&$09.15^{2.0}$	&$10.03^{3.0}$	&$11.36^{5.0}$	&$08.69^{1.0}$	&$13.15^{7.0}$	&$11.09^{4.0}$\\
CAL500	&$17.23^{7.0}$		&$01.16^{1.0}$	&$15.10^{6.0}$	&$14.50^{5.0}$	&$10.50^{4.0}$	&$03.07^{3.0}$	&$01.91^{2.0}$\\
enron	&$05.92^{6.0}$		&$04.89^{1.0}$	&$06.00^{7.0}$	&$05.80^{5.0}$	&$05.09^{2.0}$	&$05.40^{3.0}$	&$05.73^{4.0}$\\
mediamill&$03.57^{5.0}$		&$02.93^{1.0}$	&$03.50^{4.0}$	&$03.40^{3.0}$	&$03.29^{2.0}$	&$03.82^{6.0}$	&$03.95^{7.0}$\\
\midrule
Av. Rank	&$4.71$		&$1.43$	&$5.00$	&$4.71$	&$3.43$	&$4.86$	&$3.86$\\
Final Pos.	&$4.5$		&$1.0$	&$7.0$	&$4.5$	&$2.0$	&$6.0$	&$3.0$\\
\bottomrule
\end{tabular}
}
\end{table*}

Based on the \emph{accuracy} results, the average rank provides a clear indication of the studied algorithms relative
performance: \mlaslcs ranks second after \rakel and outperforms all its rivals in 3 out of the 7 studied problems, including the relatively high-complexity CAL500 problem. The comparison results are less favorable for \mlaslcs when based on the \emph{exact match} and \emph{Hamming loss} metrics, as it ranks third in both cases. Still, \mlaslcs achieves the best exact match value for 3 out of the 7 studied problems, including the CAL500 problem. In the latter case, \mlaslcs (with the \textsc{Best} inference strategy) outperforms its rivals by at least 70\%. We consider this result indicative of our proposed algorithm's ability to effectively model label correlations, given the high label cardinality ($26.04$) of the problem. 

Regarding the statistical significance of the measured differences in algorithm ranks, the use of the Friedman test does \emph{not} reject the null hypothesis (at $\alpha$=0.05) that all algorithms perform equivalently, when applied to rankings based on the accuracy and exact match metrics. %On the other hand, 
The same null hypothesis is rejected (at $\alpha$=0.05) when the studied algorithms are ranked based on Hamming loss, and the Nemenyi post-hoc test detects a significant performance difference between \rakel and (a)~HOMER and CC at $\alpha$=0.1, and (b)~ECC and BR at $\alpha$=0.05.

Overall, regardless of the evaluation metric used, \mlaslcs outperforms at least 4 of its 6 rivals. In the cases of accuracy and Hamming loss, the outperformed rivals include the state-of-the-art algorithms HOMER and CC that have been recommended as benchmarks by a recent extensive comparative study of multi-label classification algorithms \citep{madjarov2012}. Additionally, no statistically significant performance differences are detected between \mlaslcs and the best performing \rakel algorithm, with respect to all evaluation metrics. 
Thus, we consider obtained results indicative of (i) the potential of our proposed LCS approach for effective multi-label classification, as well as (ii) the flexibility of the generalized multi-label rule format that can mimic the knowledge representations induced by the studied rule-based, lazy and SVM-based ensemble learners, depending on the problem type.

\section{Conclusions and Future Work}
\label{sec:concl}
In this paper, we presented a \emph{generalized rule format} suitable for generating compact and accurate rulesets in multi-label settings. The proposed format extends existing rule representations with a flexible mechanism for modeling label correlations without the need to explicitly specify the label combinations to be considered. Thus, algorithms inducing generalized multi-label rules can approach all possible spectra between the BR (no label correlations) and LP (all possible label combinations) transformations, while producing comprehensible knowledge in the form of ``if-then'' rules.

In addition to detailing the generalized multi-label rule format, our current work also employed it in the context of a \emph{multi-label LCS algorithm}, named \mlaslcs, that is based on  a supervised LCS learning framework, properly modified to meet the new requirements posed by the multi-label classification domain. 
Its extensive experimental evaluation, missing from previous research in the area, revealed that it is capable of consistently  effective classification and highlighted it as the first LCS-based alternative to state-of-the-art multi-label classification methods. Based on the average rank over the three evaluations metrics employed, \mlaslcs came second with $2.67$ to \rakel's average first place, while it outperformed HOMER (whose average rank is $3.83$) that has recently been identified as a top-performing benchmark multi-label classification method \citep{madjarov2012}.

Regarding the combined potential of \mlaslcs and the proposed \emph{generalized multi-label rule format}, it is also worth noting that they are, with small modifications to the internal representation of rule labels, directly applicable to the relatively new task of \emph{multi-dimensional classification}.

The current limitation of our approach with respect to the arguably long times required for model training -- that is  also a problem for several non-evolutionary multi-label approaches, such as \rakel and ECC  --
 can be overcome by exploiting the parallelization, and thus scalability, potential of GAs.

An additional important issue, that needs to be addressed in future work, concerns the readability of the knowledge representations evolved, both in terms of rule quality (generalization degree) and quantity. Our first step towards this direction will be an experimental investigation of rule compaction methods available in the literature. Furthermore, based on the encouraging results obtained with the use of our clustering-based initialization procedure, alternative rule initialization methods 
will be explored, as a means to boost the predictive accuracy and interpretability of the induced knowledge representations.

\small
\section*{Acknowledgment}
The first author would like to acknowledge that this research has been funded by the Research Committee of Aristotle University of Thessaloniki, through the ``Excellence Fellowships for Postdoctoral Studies'' program.

\bibliographystyle{apalike}
\bibliography{bibliography} 

\end{document}